%% file: main.tex
\documentclass[10pt,twocolumn,letterpaper]{article}

\usepackage[pagenumbers]{cvpr} 

\definecolor{cvprblue}{rgb}{0.21,0.49,0.74}
\usepackage[pagebackref,breaklinks,colorlinks,allcolors=cvprblue]{hyperref}

\def\paperID{15865} 
\def\confName{CVPR}
\def\confYear{2026}

\usepackage{booktabs}
\usepackage{makecell}
\usepackage{lipsum}
\usepackage{graphicx}
\usepackage{xspace}
\usepackage{enumitem}
\usepackage{multirow}
\usepackage{amsmath}
\usepackage[dvipsnames,table,xcdraw]{xcolor}
\usepackage{wrapfig}
\usepackage{tabularx}
\usepackage{arydshln}
\graphicspath{ {./images/} }

\usepackage{tikz} 

\definecolor{mygrey}{HTML}{cfcfcf}   
\definecolor{mygreen}{HTML}{8de5a1}   
\definecolor{myblue}{HTML}{a1c9f4}   
\definecolor{mypurple}{HTML}{d0bbff} 
\definecolor{myred}{rgb}{1.0, 0.6, 0.6}     

\definecolor{darkgreen}{HTML}{006401}   
\definecolor{darkred}{HTML}{8B0000}   

\definecolor{dataset}{gray}{0.9}    
\definecolor{reasoningllm}{RGB}{251,221,203}
\definecolor{ours}{RGB}{218,251,203}
\definecolor{basellm}{RGB}{222,244,252}

\newcommand{\method}{\textsc{VLP}\xspace}

\newcommand{\methodfull}{\textsc{Vision-Language Programs}\xspace}

\usepackage{listings}
\definecolor{lightgray}{gray}{0.95}
\usepackage{tcolorbox}
\newtcolorbox{llmprompt}[1]{%
  colback=blue!5, colframe=blue!30, boxrule=1.5pt, width=\textwidth, 
  title=\vspace{3pt}#1, fonttitle=\bfseries,
  arc=2mm,
  fontupper=\small
}

\newcommand{\symbols}[1]{\texttt{\small{#1}}}

\makeatletter
\renewcommand\paragraph
  {%
    \@startsection{paragraph}{4}{\z@}{3.25ex \@plus 1ex \@minus .2ex}{-1ex}
      {\normalfont\normalsize\bfseries}
  }
\makeatother

\title{Synthesizing Visual Concepts as Vision-Language Programs}

\author{Antonia Wüst$^1$\\
\and
Wolfgang Stammer$^2$\\
\and
Hikaru Shindo$^1$\\
\and 
Lukas Helff$^{1,3}$\\
\and
Devendra Singh Dhami$^4$\\
\and
Kristian Kersting$^{1,3,5}$\\
\and
$^1$AIML Lab, TU Darmstadt \quad\quad
$^2$Max Planck Institute for Informatics, SIC\\
$^3$Hessian Center for AI (hessian.AI) \quad\quad
$^4$Uncertainty in AI Group, TU Eindhoven\\
$^5$German Research Center for AI (DFKI) \\
}

\begin{document}
\maketitle
\input{sec/0_abstract}    
\input{sec/1_intro}
\input{sec/acknowledgments}
{
    \small
    \bibliographystyle{style/CVPR/ieeenat_fullname}
    \bibliography{main}
}

\clearpage

\onecolumn
\appendix
\input{sec/X_suppl}

\end{document}

%% file: sec/0_abstract.tex
\begin{abstract}
Vision-Language models (VLMs) achieve strong performance on multimodal tasks but often fail at systematic visual reasoning tasks, leading to inconsistent or illogical outputs. Neuro-symbolic methods promise to address this by inducing interpretable logical rules, though they exploit rigid, domain-specific perception modules. We propose Vision-Language Programs (VLP), which combine the perceptual flexibility of VLMs with systematic reasoning of program synthesis. Rather than embedding reasoning inside the VLM, VLP leverages the model to produce structured visual descriptions that are compiled into neuro-symbolic programs. The resulting programs execute directly on images, remain consistent with task constraints, and provide human-interpretable explanations that enable easy shortcut mitigation. Experiments on synthetic and real-world datasets demonstrate that VLPs outperform direct and structured prompting, particularly on tasks requiring complex logical reasoning. \footnote{Project page: \href{https://ml-research.github.io/vision-language-programs/}{ml-research.github.io/vision-language-programs}}
\end{abstract}

%% file: sec/1_intro.tex


\section{Introduction}


\begin{figure}[h]
    \centering
    \includegraphics[width=0.99\linewidth]{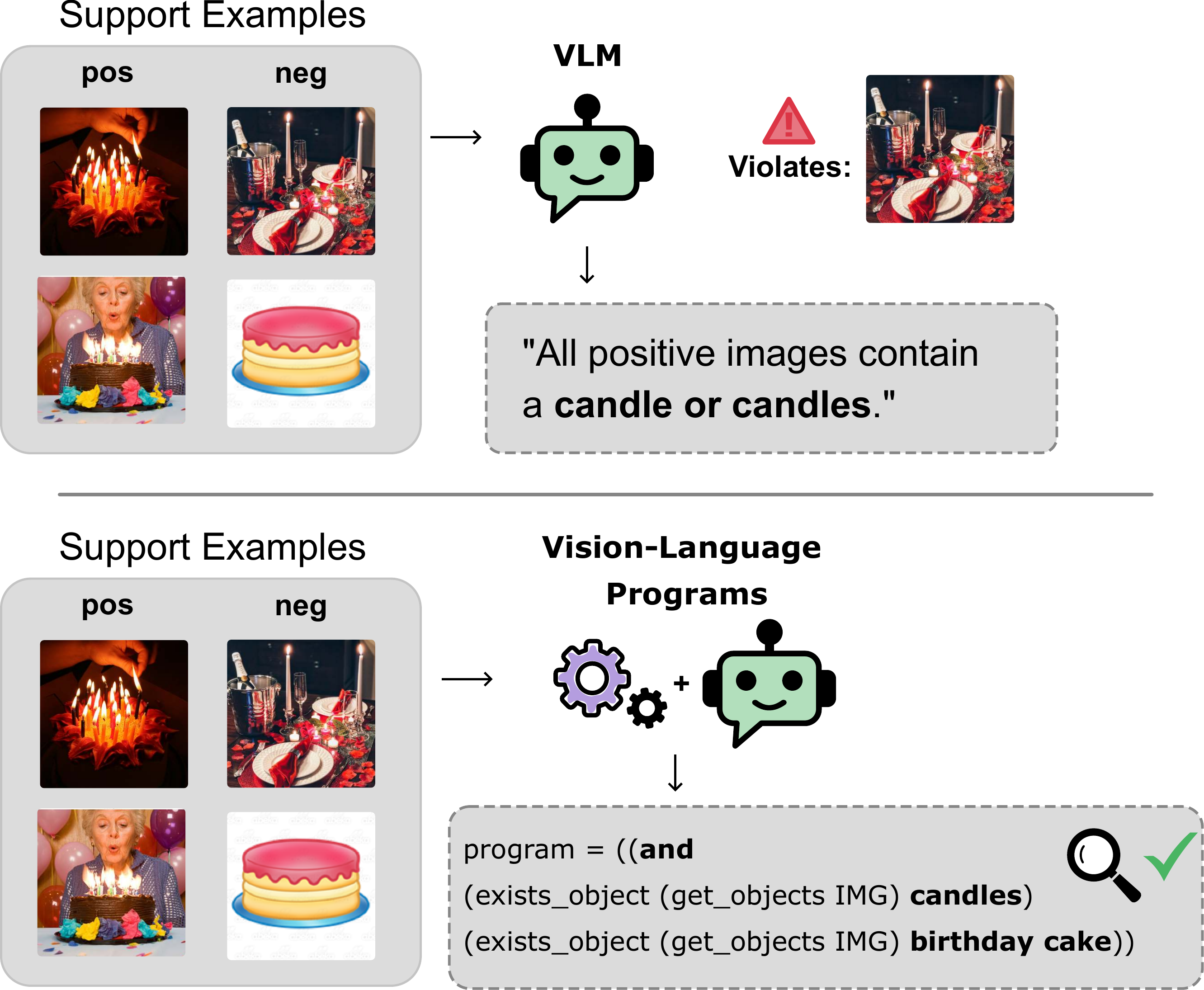}
    \caption{
    \textbf{VLMs cannot reliably perform inductive logic learning from images}, failing to capture visual compositions like \emph{``candles~\textbf{and} birthday cake''}.
    \textbf{Vision-Language Programs (VLP)} employ explicit symbolic reasoning to overcome such visual reasoning errors while maintaining perceptual flexibility.
    }
    \label{fig:motivation}
\end{figure}

Vision–language models (VLMs) have achieved impressive results across multimodal tasks, yet they continue to struggle with visual reasoning. Studies reveal frequent failures in both perception and reasoning, even on relatively simple tasks \citep{wustbongard, kamath2023whatsup, rahmanzadehgervi2024vision, zhou2023rome, zhang2024far, wang2024exploring, grounded_rl_vr, helff2024vlol}. \Eg, in inductive visual reasoning tasks where models must propose rules that distinguish between image sets (\cf \autoref{fig:motivation}), VLMs often fail by generating statements that violate the task constraints. In this example, the VLM proposes the rule "contains candle or candles", incorrectly satisfying one of the negative images. Such errors highlight the gap between pattern recognition and systematic reasoning in VLMs.

Recent work attempts to address this gap through test-time scaling, where models “think” longer via extended chain-of-thought generation~\citep{wei2022chain, snell2024scaling}. While effective in some cases, this approach is computationally expensive and prone to contradictions or repetitive loops \citep{shojaee2025illusion,zeng2025revisiting, wu2025s, helff2025slr}. This raises the question of whether language-based inference alone is sufficient for robust reasoning.

Neuro-symbolic (NeSy) AI offers a promising alternative by integrating neural processing with structured symbolic inference \citep{garcez2015neural, sarker2021neuro, kautz2022third, manhaeve2018deepproblog, shindo23alphailp, shindo23neumann, wustpix2code}. This paradigm has demonstrated improved robustness, compositional generalization, and interpretability, \eg, over monolithic VLMs under domain shifts \citep{kamali2025neptune}. Hereby, program synthesis \citep{gulwani2017program}, which induces interpretable and logically consistent programs from examples, provides a particularly natural mechanism for implementing this integration in visual reasoning tasks.
However, existing neuro-symbolic approaches for visual reasoning face critical limitations. Methods either require explicit queries to drive program generation, limiting their applicability to inductive reasoning tasks, or depend on domain-specific image encoders, preventing generalization across diverse visual domains.

We therefore propose combining VLMs with program synthesis in form of \methodfull (\method) to overcome these shortcomings. 
Importantly, instead of embedding reasoning inside the VLM, VLP produces structured visual descriptions that can be compiled into symbolic programs, decoupling perception from reasoning. \method automatically induces symbolic rules from small sets of labeled image examples by first discovering candidate symbols through VLM-based analysis, then defining VLM functions that extract structured representations from images, which can be composed with symbolic reasoning functions in a domain-specific language. This allows the resulting programs to leverage neural perception at execution time while maintaining symbolic interpretability and logical consistency. This dualistic process allows \method to \textit{execute} directly on images.

Evaluations on both synthetic and real-world datasets demonstrate that even small VLMs, when embedded in our framework, surpass direct prompting, especially on tasks requiring complex logical reasoning. This hybrid approach thus leverages the perceptual priors of VLMs while enabling symbolic reasoning, marking a crucial step toward models that not only achieve strong performance but also provide transparent, structured decision processes.

To this end, we introduce: (i) \methodfull, a framework combining VLMs with program synthesis to induce symbolic rules from labeled images without hand-crafted detectors or task-specific queries; (ii) a domain-specific language integrating compositional VLM perception functions with symbolic reasoning operators; (iii) a probabilistic synthesis procedure that discovers and ranks programs by accuracy and likelihood; (iv) comprehensive empirical evidence showing our approach outperforms direct prompting, particularly on logically complex tasks; and (v) analysis demonstrating how programmatic structure enables shortcut detection and mitigation through transparent decision processes.


\section{Related Work}

\textbf{Neuro-Symbolic Concept Induction.}
Neuro-symbolic AI~\citep{garcez2015neural,sarker2021neuro,kautz2022third} seeks to integrate the strengths of neural representations with the structure of symbolic reasoning. A common approach is to extract structured representations from raw perceptual inputs, then perform rule learning over the resulting symbols~\citep{Yang2020NLIL,shindo23alphailp,shindo23neumann,sunder_one_shot_ie_2019,wustpix2code,stammer2021right}. These methods typically extract structured representations from raw inputs, then perform rule learning over the resulting symbols, for instance via inductive logic programming (ILP) \citep{shindo23alphailp,shindo23neumann} or probabilistic logic frameworks \citep{manhaeve2018deepproblog,skryagin2021slash,skryagin2023scalable}. While effective for complex synthetic scenes, their applicability to open-world settings is often limited due to reliance on pre-defined predicates and domain-specific object detectors for constructing intermediate symbolic representations.


\textbf{Program-based Visual Reasoning.}
A prominent recent approach to complex visual reasoning leverages programs for systematic image analysis. Frameworks such as VisProg~\citep{VisProg}, ViperGPT~\citep{ViperGPT}, CodeVQA~\citep{CodeVQA}, and NePTune~\citep{kamali2025neptune} employ foundation models to generate executable programs conditioned on visual inputs and natural language instructions or questions. In contrast, our approach (\method) focuses on inducing programs from labeled visual examples in the absence of task-specific queries, thereby uncovering programmatic representations that explain the conceptual differences between them.
\textbf{Program synthesis}~\citep{gulwani2011automating} has long been successful in rule induction, with early efforts focusing on domains such as list processing and text editing~\citep{gulwani2017program, ellis2023dreamcoder}. A benefit of explicit program synthesis over LLM-prompted approaches is the guarantee of syntactically valid and executable programs, eliminating formatting errors that can hinder downstream reasoning. More recent work extends these ideas to abstract reasoning tasks, such as the Abstract Reasoning Corpus~\citep{chollet2019measure}, by leveraging foundation models to discover higher-level concepts~\citep{wang2024hypothesis, barke2024hysynth}. However, the intersection of program synthesis and foundation models for visual rule induction remains largely unexplored.
An initial step in this direction was made by~\citet{wustpix2code}, who proposed a program synthesis approach for visual concept induction, inducing programs from positive and negative visual samples. Their method, however, depends on domain-specific object detectors to convert images into symbolic representations, which limits its generality. In contrast, \method~extends this idea by performing program synthesis directly on natural images, eliminating the need for domain-specific pretraining and enabling broader applicability. 

\section{Vision Language Programs}

\begin{figure*}[t!]
    \centering
    \includegraphics[width=\linewidth]{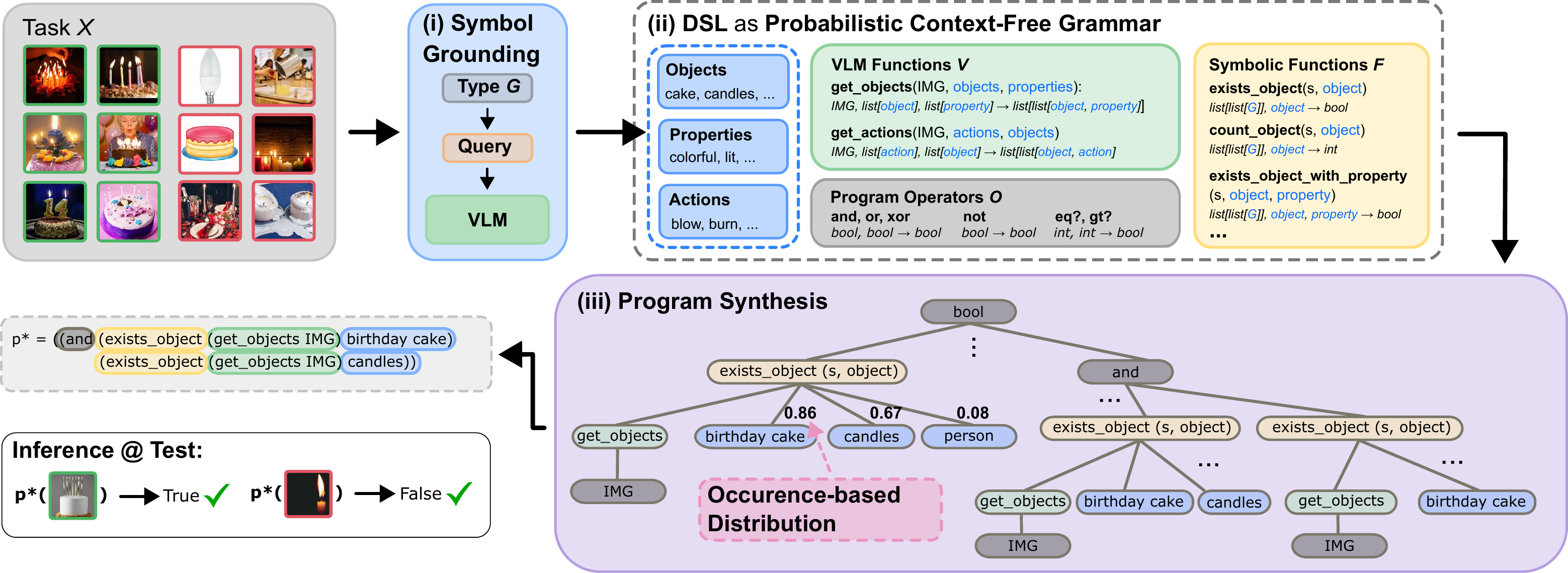}
    \caption{\textbf{Overview of \methodfull synthesis}. Relevant variables are first discovered from the input examples (i) and used to construct a task-specific DSL, including VLM-based functions (ii). Program synthesis (iii) then searches this space to retrieve the most probable program that also achieves the highest accuracy on the input.}
    \label{fig:method}
\end{figure*}

We introduce \methodfull (\method), a framework designed to induce symbolic programs that explain the underlying visual rule from a small set of labeled image examples. \method combines the perceptual strengths of VLMs with structured reasoning in a Domain-Specific Language (DSL), allowing for explicit, interpretable, and systematic reasoning grounded in visual perception.

\method operates in three consecutive stages. First, during \emph{symbol grounding (i)}, relevant object, property, and action symbols are grounded for the task at hand. Second, a \textit{Probabilistic Context-Free Grammar} is formed from the Domain-Specific Language (DSL) and the previously ground symbols, including symbolic and neural, VLM-based functions. This forms a probabilistic interface for program search based on visual inputs. Finally, \emph{program synthesis (iii)} searches over the solution space created by the grammar to synthesize a program that best distinguishes positive from negative examples. The resulting output program captures the semantic relations between the input images and can be executed on new test examples to infer new labels. We next detail each stage.

\subsection{Problem Setup}
We formulate inductive visual reasoning as the task of discovering a latent visual rule that explains a set of example images (denoted as \emph{few-shot} examples in the remainder).
Formally, let
$\mathcal{X} = \{(I_1, y_1), (I_2, y_2), \ldots, (I_n, y_n)\}$
denote a task, where each $I_i$ is an input image and $y_i \in \{0, 1\}$ is the corresponding binary label. A label $y_i = 1$ indicates that $I_i$ satisfies the latent visual rule (positive example), whereas $y_i = 0$ indicates that it does not (negative example). Each task is additionally associated with a set of held-out \emph{query} samples for evaluation.

\subsection{Symbol Grounding}\label{sec:variable}
The first stage of \method establishes an interface between continuous visual inputs and discrete symbolic representations. This process, which we refer to as \emph{symbol grounding}, maps perceptual information from images into structured, type-constrained symbols that form the atomic units for subsequent reasoning. We define three fundamental symbol types
$$G = \{\symbols{object}, \symbols{property}, \symbols{action}\}$$
These types constrain the semantic roles that individual symbols play in downstream program construction.
Rather than relying on a fixed vocabulary, the vocabulary is dynamically adjusted based on the task at hand. This preserves generality and adaptation to novel domains and unseen visual compositions. In detail, given a reasoning task $\mathcal{X}$, \method provides task-specific groundings of these abstract types by querying a pretrained VLM $\mathcal{M}$. For each symbol type $G_i \in G$, the model proposes a set of groundings $E_i$:
\begin{equation}
\mathcal{M}(G_i, \mathcal{X}) = E_i, \quad E_i = \{e_1, \ldots, e_m\},
\end{equation}   
where each $e \in E_i$ denotes an individual grounding for symbol type $G_i$. E.g., in \autoref{fig:method}, \symbols{cake} and \symbols{candles} ground symbol type \symbols{object}, while \symbols{colorful}, \symbols{blow}, and \symbols{burn} ground \symbols{property} and \symbols{action} symbols, respectively. The process is guided by type-specific queries (\cf~\autoref{app:variable_discovery_prompts}).

The result is a structured pool of symbols $\mathcal{E} := \{E_1, \ldots, E_{|G|}\}$ that captures the objects, properties, and actions relevant to the current task. This pool acts as the semantic substrate for forming visual concepts.


\subsection{Vision-Language DSL}
The second element of \method is a Domain-Specific Language (DSL) \citep{fijalkow2022scaling} that formalizes an interface between perception and reasoning. Unlike task-specific DSLs, ours is VLP-specific: it defines a general symbolic interface that remains invariant across domains and tasks. While the grounded symbols capture task-specific semantics, the DSL defines the syntax and functional structures shared across all tasks. It includes a small set of syntactic primitives such as \symbols{bool} and \symbols{int}, and a dedicated type \symbols{image}, representing the input domain. Beyond these primitives, the DSL defines three function types: \emph{VLM functions}, \emph{symbolic functions}, and \emph{program operators}. VLM functions map perceptual features from images to symbolic representations, symbolic functions encode logical or arithmetic operations over these symbolic representations, and program operators compose these components into executable reasoning programs. 

\paragraph{VLM functions $\mathcal{V}$} equip \method with a perceptual interface to extract symbolic information from visual inputs using VLMs. Each function ($v \in \mathcal{V}$) takes as input an image ($I$) together with one or more sets of ground symbols ($E_i$) obtained during symbol grounding (\autoref{sec:variable}), and outputs a nested symbolic representation $s$:
\begin{equation}
v(I, E_1, \ldots, E_m) = s,
\end{equation} 
where $E_i \in \texttt{list}[G_i]$ with $G_i \in \mathcal{G}$ for $m \geq 1$, and $s \in \texttt{list}[\texttt{list}[\mathcal{G}]]$. In essence, VLM functions translate raw visual inputs into structured symbolic representations. For example, the VLM function \symbols{get\_objects} (see \autoref{fig:method}) takes an image along with the symbol groundings for \textit{object} and \textit{property}, and extracts a structured \textit{object–property} mapping for the given image. For the first input image (top left in \autoref{fig:method}), this mapping could be represented as:
\[
\symbols{[[birthday cake], [candles, colorful]]}.
\]
The output provides a symbolic representation of an image’s semantics for downstream reasoning. Since $\mathcal{E}$ is constant across task images, we omit explicit symbol inputs and write functions like \symbols{get\_objects} and \symbols{get\_actions} as depending only on the image $I$ in the following.

\paragraph{Symbolic functions $\mathcal{F}$} form the core reasoning primitives of \method. They operate directly on the symbolic representations $s$ and capture the relationships, attributes, and interactions among them. Each function represents an interpretable reasoning step.
For instance, the function \symbols{exists\_object(s, e)} in \autoref{fig:method} evaluates whether an object $e$ appears in the representation $s$, returning a boolean. Other symbolic functions count objects or properties, verify the presence of specific actions, or symbol combinations. Collectively, these functions form the reasoning vocabulary of \method, enabling structured queries over visual elements.

\paragraph{Program operators $\mathcal{O}$} specify how reasoning primitives in \method can be composed. They encompass logical connectives (\symbols{AND}, \symbols{OR}, \symbols{NOT}) and comparison operators ($=, <, >$), enabling the construction of complex executable programs that represent abstract visual concepts.

\subsection{DSL to Probabilistic Context-Free Grammar}
With semantics defined by \textit{symbol grounding} and syntax specified by the \textit{DSL}, \method employs a Probabilistic Context-Free Grammar (PCFG) \citep{fijalkow2022scaling} to formalize program synthesis.
The PCFG serves as a prior over the space of syntactically valid programs, defining probabilistic production rules from the DSL that guide program search.


Formally, a PCFG is defined as a tuple $G = (N, T, R, S, P)$, where $N$ denotes nonterminal symbols (types), $T$ denotes terminal symbols (grounded symbols), $R$ is the set of production rules, $S$ is the start symbol, and $P$ assigns probabilities to each rule in $R$. The set of all possible symbol types is given by:
\[
\mathcal{T} = {G \cup \{\symbols{image}, \symbols{bool}, \symbols{int}\}},
\]
Each type $\tau \in \mathcal{T}$ corresponds to a nonterminal in $N$, representing all expressions that evaluate to that type.

\paragraph{Grammar Construction}
The core of the PCFG lies in the production rules $R$ that govern how programs are composed from DSL primitives and grounded symbols. 
Each type corresponds to one or more productions rules that define which expressions from the DSL lead to this type.
A general production rule $r\in R$ takes the form:
\[
r: \tau_i \rightarrow f(\tau_1, \tau_2, \ldots, \tau_k),
\]
where $f \in \mathcal{V} \cup \mathcal{F} \cup \mathcal{O} \cup \mathcal{E}$ is a DSL element producing an output of type $\tau_i$, and each $\tau_j$ denotes the expected argument type. Applying these rules recursively expands the start symbol $S$ (of output type \texttt{bool}) until all nonterminals are resolved, yielding a complete, type-consistent program.

Enumerating all such rules under $\mathcal{T}$ defines the complete space of syntactically valid, type-consistent programs under $R$. For example, several valid rules for constructing boolean expressions would be:
\[
\small
\begin{aligned}
\symbols{bool} &\;\rightarrow\; \symbols{\textbf{and}(bool, bool)} \\
\texttt{bool} &\;\rightarrow\; \texttt{\textbf{or}(bool, bool)} \\
\texttt{bool} &\;\rightarrow\; \texttt{\textbf{exists\_object}(s, object)}.
\end{aligned}
\]
Likewise, perceptual rules may link to visual grounding functions, such as:
\[
\small
\begin{aligned}
\texttt{s} &\;\rightarrow\; \texttt{\textbf{get\_objects}(IMG)} \\
\symbols{object} &\;\rightarrow\; \symbols{\textbf{birthday cake}}
\end{aligned}
\]
Repeatedly applying such rules expands \symbols{bool} into a complete, type-consistent program operating on visual inputs.

\paragraph{Probabilistic Weighting.} 
Each production rule $r \in R$ is associated with a probability $P(r)$ that determines how likely specific functions, operators, or grounded symbols $\mathcal{E}$ are to be selected during synthesis. Since \method is training-free, most rules are assigned uniform probabilities. The only exception concerns the symbols $\mathcal{E}$ discovered by the VLM $\mathcal{M}$. Here we leverage their relative occurrence frequencies in positive and negative examples for weighting their likelihood. This employs a simple yet effective form of data-driven inductive bias.
We estimate these weights using an occurrence-based distribution, where a symbol $e \in \mathcal{E}$ that occurs more frequently in positive examples is assigned higher probability:
\begin{equation}
P(e) \propto
\frac{n_{\text{pos}}}{N_{\text{pos}}} \cdot \frac{n_{\text{pos}}}{n_{\text{pos}} + n_{\text{neg}}},
\end{equation}
where ($n_{\text{pos}}$) and ($n_{\text{neg}}$) denote the number of positive and negative occurrences respectively, and ($N_{\text{pos}}$) is the total count of positive examples. When ($n_{\text{pos}} = 0$), a constant ($\epsilon = 0.01$) is used to avoid zero-probability assignments.


\subsection{Program Synthesis}  
Given the PCFG, \method performs program synthesis by searching for an executable program $p$ that best explains the task $\mathcal{X}$. Each program transforms an input image $I_i$ into a boolean prediction $\hat{y}_i = p(I_i)$, whose correctness is evaluated against the ground-truth label $y_i$.

\paragraph{Program search.}  
The search explores the space of candidate programs defined by the grammar. For each candidate $p$, we compute its accuracy on $\mathcal{X}$,
\[
\text{Acc}(p) = \frac{1}{n} \sum_{i=1}^n \mathbf{1}\big[p(I_i) = y_i\big],
\]
and its probability under the PCFG, $P(p)$, given by the product of the probabilities of the rules used to construct $p$. To improve efficiency, outputs of all VLM functions $\mathcal{V}$ are precomputed for every $I_i \in \mathcal{X}$.

Candidate programs are ranked by a two-level criterion:  
1. \emph{Primary:} accuracy $\text{Acc}(p)$;  
2. \emph{Secondary:} probability $P(p)$ to break ties.  
The top-ranked program $p^*$ is selected as the final solution (\cf \autoref{fig:method}, bottom left).

\begin{table*}[!t]
    \centering
    \caption{\textbf{Comparison of base VLMs and VLMs with VLP (averaged over three runs).} Balanced accuracy (\%) on 6-shot Bongard tasks and 10-shot logical reasoning benchmarks. Improvements (\textcolor{darkgreen}{green} / \textcolor{gray}{gray}) denote changes relative to the baseline. Best results per model are shown in \textbf{bold}; overall best results per dataset column are \underline{underlined}.}
    \label{tab:vlp_results}
    \vskip 0.4em
    \small
    \resizebox{0.98\linewidth}{!}{
    \begin{tabular}{lllllll}
    \toprule
    \textbf{Model} & \textbf{Avg.} &
    \multicolumn{3}{c}{\textbf{Bongard Tasks (6-shot)}} &
    \multicolumn{2}{c}{\textbf{Logical Reasoning (10-shot)}} \\
    \cmidrule(lr){3-5} \cmidrule(lr){6-7}
     &  & Bongard-HOI & Bongard-OW & Bongard-RWR & COCOLogic & CLEVR-Hans3 \\
    \midrule
    InternVL3-8B & $57.4$ & $60.5$ & $59.2$ & $47.2$ & $71.5$ & $48.3$ \\
    \rowcolor{gray!10} \quad w/ \textbf{VLP} &
        \textbf{\underline{70.9}} {\footnotesize\textcolor{darkgreen}{(+13.5)}} &
        \textbf{\underline{77.7}} {\footnotesize\textcolor{darkgreen}{(+17.2)}} &
        \textbf{67.5} {\footnotesize\textcolor{darkgreen}{(+8.3)}} &
        \textbf{53.9} {\footnotesize\textcolor{darkgreen}{(+6.7)}} &
        \textbf{\underline{81.0}} {\footnotesize\textcolor{darkgreen}{(+9.5)}} &
        \textbf{74.4} {\footnotesize\textcolor{darkgreen}{(+26.1)}} \\
        \midrule
    InternVL3-14B & $61.5$ & $66.9$ & $62.5$ & $51.4$ & \textbf{78.4} & $48.3$ \\
    \rowcolor{gray!10} \quad w/ \textbf{VLP} &
        \textbf{68.3} {\footnotesize\textcolor{darkgreen}{(+6.8)}} &
        \textbf{74.3} {\footnotesize\textcolor{darkgreen}{(+7.4)}} &
        \textbf{64.7} {\footnotesize\textcolor{darkgreen}{(+2.2)}} &
        \textbf{52.8} {\footnotesize\textcolor{darkgreen}{(+1.4)}} &
        $76.7$ {\footnotesize\textcolor{gray}{(-1.7)}} &
        \textbf{73.3} {\footnotesize\textcolor{darkgreen}{(+25.0)}} \\
        \midrule
    Kimi-VL-A3B-Instruct & $58.5$ & $59.8$ & $58.6$ & $46.4$ & \textbf{77.9} & $50.0$ \\
    \rowcolor{gray!10} \quad w/ \textbf{VLP} &
        \textbf{65.5} {\footnotesize\textcolor{darkgreen}{(+7.0)}} &
        \textbf{69.4} {\footnotesize\textcolor{darkgreen}{(+9.6)}} &
        \textbf{59.4} {\footnotesize\textcolor{darkgreen}{(+0.8)}} &
        \textbf{52.5} {\footnotesize\textcolor{darkgreen}{(+6.1)}} &
        $70.1$ {\footnotesize\textcolor{gray}{(-7.8)}} &
        \textbf{76.1} {\footnotesize\textcolor{darkgreen}{(+26.1)}} \\
        \midrule
    Qwen2.5-VL-7B-Instruct & $60.1$ & $65.2$ & $\textbf{66.2}$ & \textbf{49.7} & $73.2$ & $46.1$ \\
    \rowcolor{gray!10} \quad w/ \textbf{VLP} &
        \textbf{69.5} {\footnotesize\textcolor{darkgreen}{(+9.4)}} &
        \textbf{68.8} {\footnotesize\textcolor{darkgreen}{(+3.6)}} &
        $62.9$ {\footnotesize\textcolor{gray}{(-3.3)}} &
        $49.2$ {\footnotesize\textcolor{gray}{(-0.5)}} &
        \textbf{80.5} {\footnotesize\textcolor{darkgreen}{(+7.3)}} &
        \textbf{\underline{86.1}} {\footnotesize\textcolor{darkgreen}{(+40.0)}} \\
        \midrule
     Qwen3-VL-30B-A3B-Instruct & $63.4$  & $69.0$ & \textbf{\underline{68.5}} & $55.8$ & $73.9$ & $50.0$  \\
    \rowcolor{gray!10} \quad w/ \textbf{VLP} &
        \textbf{68.9} {\footnotesize\textcolor{darkgreen}{(+5.5)}} &
        \textbf{74.5} {\footnotesize\textcolor{darkgreen}{(+5.5)}} &
        $66.3$ {\footnotesize\textcolor{gray}{(-2.2)}} &
        \textbf{\underline{58.3}} {\footnotesize\textcolor{darkgreen}{(+2.5)}} &
        \textbf{79.1} {\footnotesize\textcolor{darkgreen}{(+5.2)}} &
        $\textbf{66.1}$ {\footnotesize\textcolor{darkgreen}{(+16.1)}} \\
    \bottomrule
    \end{tabular}
    }
\end{table*}

\section{Experiments}

In the following, we evaluate \methodfull on several datasets and tasks to assess neuro-symbolic \method across various visual reasoning settings. We hereby address the following research questions:

\begin{itemize}
    \item Can \method improve concept learning performance over VLMs across varying visual domains? (\textbf{RQ1})
    \item Can VLPs with non-reasoning models improve over dedicated reasoning models? (\textbf{RQ2})
    \item Do VLP models exhibit advantages over VLMs with an increased number of training samples? (\textbf{RQ3})
    \item Do \method facilitate knowledge incorporation, \eg, for shortcut mitigation? (\textbf{RQ4})
\end{itemize}


\noindent \textbf{Data.}
For the evaluation of our method, we use a set of different datasets across multiple domains that are based on synthetic and real-world images. 
Specifically, we evaluate on the datasets Bongard-HOI \citep{jiang2022bongard}, Bongard-OpenWorld \citep{wubongard} and Bongard-RWR\citep{malkinskireasoning}, which are based on real-world images and incorporate a diverse set of visual concepts. For a real-world dataset that provides more complex logical rules, we utilize COCOLogic \citep{steinmann2025object} and create $10$ tasks from it, one for each class. For the synthetic dataset, we use CLEVR-Hans3 \citep{stammer2021right}, where we leverage the three classes to construct three tasks with complex logical rules from it. The Bongard datasets provide $12$ samples from which the target rule should be induced, $6$ positives and $6$ negatives. For the other two datasets, we take $20$ balanced support samples. Additional details are provided in \autoref{app:data}.

\noindent \textbf{Experimental Setup.} We incorporate and compare to a selection of open source VLMs with varying sizes. Namely, we utilize InternVL3-8B and InternVL3-14B \citep{chen2024internvl}, Kimi-VL-A3B-Instruct \citep{kimiteam2025kimivltechnicalreport}, as well as Qwen2.5-VL-72B \citep{qwen2.5-VL} and 
Qwen3-VL-30B-A3B-Instruct \cite{qwen3technicalreport}. 
We prompt a model three times using sampling-based generation and perform program synthesis using the Heap Search Algorithm from \citep{fijalkow2022scaling} with a $10$-second budget. The maximum program depth is $4$ for Bongard-OpenWorld and Bongard HOI, and $6$ for COCOLogic and CLEVR-Hans3. The DSL used for the experiments is provided in \autoref{sec:dsl}.
In the context of \textbf{RQ1} we compare the base models with and without VLP integration. 
In the context of \textbf{RQ2} we compare instruction-tuned models with VLP to dedicated reasoning models Kimi-VL-A3B-Thinking \citep{kimiteam2025kimivltechnicalreport}, Qwen3-VL-30B-A3B-Thinking \cite{qwen3technicalreport} and gpt-5 \cite{gpt-5}.
For \textbf{RQ3}, we evaluate all models from RQ1 using an increased number of support samples and average their results. 
For evaluations regarding knowledge incorporation (\textbf{RQ4}) we use CLEVR-Hans3 to investigate how DSL edits can be used to improve performance and mitigate shortcut learning. 
Across all evaluations, model performance is measured using balanced accuracy, reflecting each model’s ability to correctly classify the query (test) images. 

\paragraph{VLPs boost VLMs across domains (RQ1).}
In our first evaluation, we investigate the potential of our neuro-symbolic VLP framework to leverage the power of VLMs in diverse visual concept learning tasks.
For this, we compare the performance of five different base VLMs with and without VLP processing on five datasets. 

We observe in \autoref{tab:vlp_results} that VLPs obtain substantially higher results across all models, with improvements of up to $13.5\%$ on average across all datasets. Interestingly, model size appears to have minimal effect on VLP performance; however, smaller models (\eg, InternVL3-8B and Qwen2.5-VL-7B) tend to benefit most from VLP-based processing. The strongest improvements across all models occur on the compositionally complex CLEVR-Hans3 dataset, where the synthetic images are likely more out-of-distribution for pretrained VLMs, usually including $5$ to $10$ objects. This pattern suggests that structured reasoning offers greater advantages when perceptual uncertainty is higher. Notably, none of the model encoders were specifically finetuned on these datasets, demonstrating that \method grants domain-independent flexibility.

\begin{figure}[t!]
    \centering
    \includegraphics[width=0.95\linewidth]{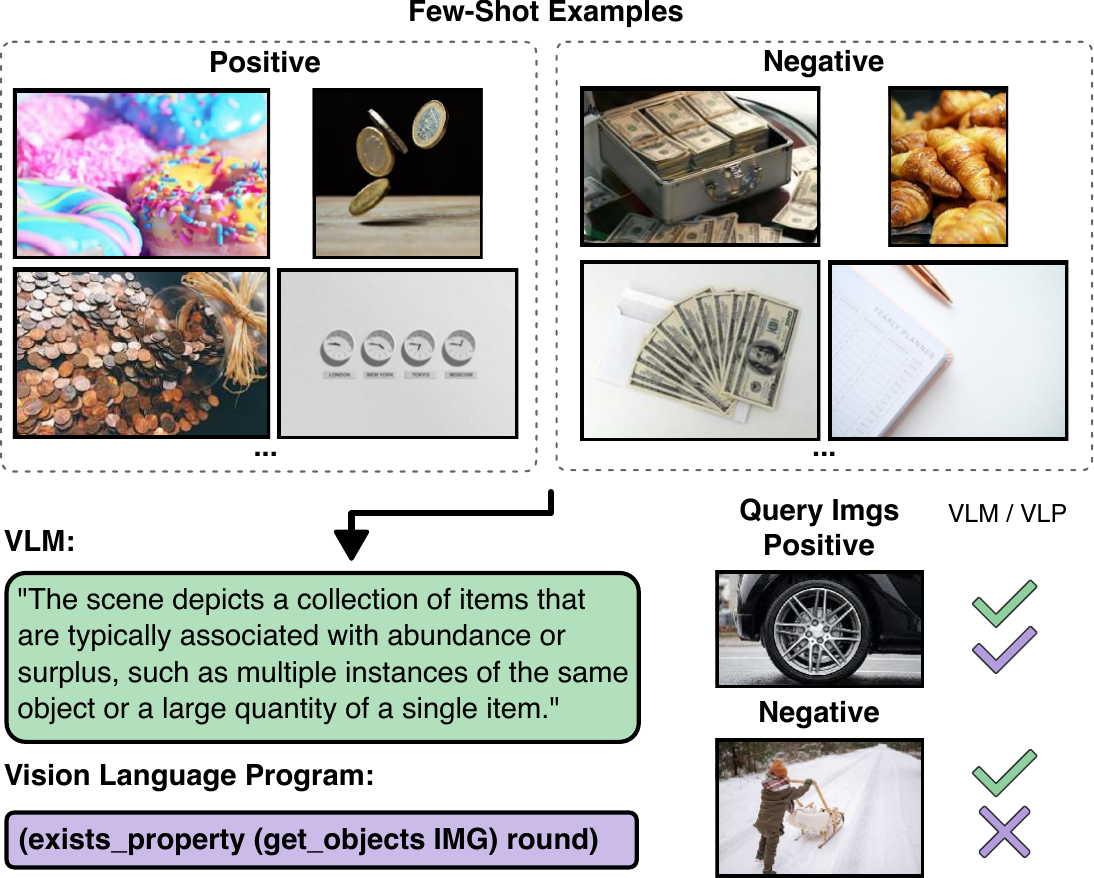}
    \caption{\textbf{Qualitative comparison on Bongard-RWR.} Direct VLM (Qwen3) prompting produces an incorrect rule about "abundance", misclassifying a query image. Qwen3 w/ \method discovers a correct program that identifies round objects and achieves perfect query classification accuracy.}
    \label{fig:pos_vlp_example}
\end{figure}

\autoref{fig:pos_vlp_example} illustrates this effect of \method's decoupling of perception from reasoning on a Bongard-RWR task. When prompted directly, the base Qwen3-VL model proposes a complex rule but fails to identify the simple underlying concept of "round vs. non-round objects," ultimately classifying both test images as positive. In contrast, the same base model with \method successfully discovers the program 
\[\footnotesize \texttt{$p^*$= (exists\_property (get\_objects IMG) \textbf{round})}\]
by first identifying individual objects in each image, then inferring through program search that all positive images contain objects with the property \texttt{round}. The resulting Vision-Language Program $p^*$ correctly classifies all test images.

The only dataset where VLP does not achieve the overall best performance is Bongard-OpenWorld. Upon examining the failure cases on this dataset, we identified numerous annotation errors (\cf \autoref{app:bongard-ow-dataset-issues}). In such cases, standard VLM prompting tends to generate plausible but loosely defined rules that accommodate these inconsistencies. In contrast, \method’s structured approach induces programs that remain logically consistent with the (incorrectly labeled) few-shot examples, but consequently fail the query samples.



Within the context of RQ1, we additionally investigate whether the observed improvements arise from the use of structured symbolic image representations (\eg,~\symbols{get\_objects}, \symbols{get\_actions}) by comparing LLM-based reasoning on symbolic inputs with VLP's reasoning in Suppl.~\autoref{app:reasoning_ablation}. The results indicate that, in fact, VLP's performance gains are primarily driven by the symbolic search process rather than the representation format.
Conclusively, the induced rules by \method are more reliable than rules obtained by prompting the model directly, even with structured symbolic representations at hand.

In summary, across these evaluations, we conclude that our neuro-symbolic VLP framework consistently enhances the reasoning capabilities of VLMs by effectively combining visual perception with symbolic search.

\begin{table}[]
    \centering
    \caption{\textbf{Comparison between \textit{thinking} and \textit{non-thinking} models with VLP}. VLP performs comparably or better than thinking models while using significantly fewer tokens, and can even build upon thinking-based approaches for further improvements.}
    \label{tab:thinking}
    \resizebox{\linewidth}{!}{
    \begin{tabular}{lrrrr}
        \toprule
          & \multicolumn{2}{c}{COCOLogic} & \multicolumn{2}{c}{CLEVR-Hans3} \\
          &  Acc & Tokens & Acc & Tokens\\
         \midrule        
         {Kimi w/ Think} & $52.2$ & $106,446$ & $30.0$ & $46,941$\\
        \rowcolor{gray!10} \small{Kimi w/ \textbf{VLP}} & $\textbf{68.3}$ & $\textbf{43,958}$ & $\textbf{83.3}$ & $\textbf{6,584}$ \\
        \midrule
        {Qwen3 w/ Think} & $\textbf{81.8}$ & 108,346 & $46.7$ & 106,922 \\
        \rowcolor{gray!10} {Qwen3 w/ \textbf{VLP}} & $78.8$ & $\textbf{22,341}$ & $\textbf{68.3}$ & $\textbf{5,052}$ \\
        
        \midrule \midrule
        
        GPT-5 w/ Think & $78.5$ & $115,813$ & $65.0$ & $98,046$ \\   
        \rowcolor{gray!10} GPT-5 w/ \textbf{VLP} & $81.8$ & $\textbf{17,404}$ & $68.3$ & $\textbf{8,058}$ \\ 
        \rowcolor{gray!10} {GPT-5 w/ Think+\textbf{VLP}} & $\textbf{84.3}$ & $183,387$ & $\textbf{70.0}$ & $84,642$  \\ 
         \bottomrule
    \end{tabular}
    }
\end{table}

\paragraph{Symbolic reasoning of VLPs outperforms VLM-based reasoning (RQ2).} 
In our next evaluation, we compare the performance of \method with instruction-tuned base VLMs to dedicated ``reasoning'' models. We focus on COCOLogic and CLEVR-Hans3 for this comparison, as these datasets exhibit the highest compositional and logical complexity, making them ideal testbeds for distinguishing structured symbolic reasoning from pure language-based inference.
We compare Kimi-VL-A3B and Qwen3-VL-30B-Instruct with \method to their ``Thinking'' counterparts. Additionally, we evaluate gpt-5 (high reasoning effort) against gpt-5-chat (no reasoning) and gpt-5 (low reasoning effort), both with \method. We conduct the experiments with one run per model. 

The results in \autoref{tab:thinking} show that integrating \method with non-reasoning models yields substantial improvements over dedicated reasoning models in all cases except Qwen3 on COCOLogic. Moreover, VLP-based reasoning requires substantially fewer tokens than dedicated reasoning models, despite executing more prompts, thereby demonstrating greater computational efficiency. Interestingly, integrating gpt-5 with low reasoning effort into \method achieves improvements over both gpt-5 with Thinking and gpt-5-chat with \method, while increasing token count only moderately on COCOLogic and even reducing it on CLEVR-Hans3. This suggests that \method reasoning provides an orthogonal enhancement to the thinking mode of VLMs, yielding complementary performance benefits.

Overall, our findings suggest that VLP achieves strong reasoning performance while maintaining computational efficiency. We thus answer RQ2 affirmatively.

\begin{figure}
    \centering
    \includegraphics[width=\linewidth]{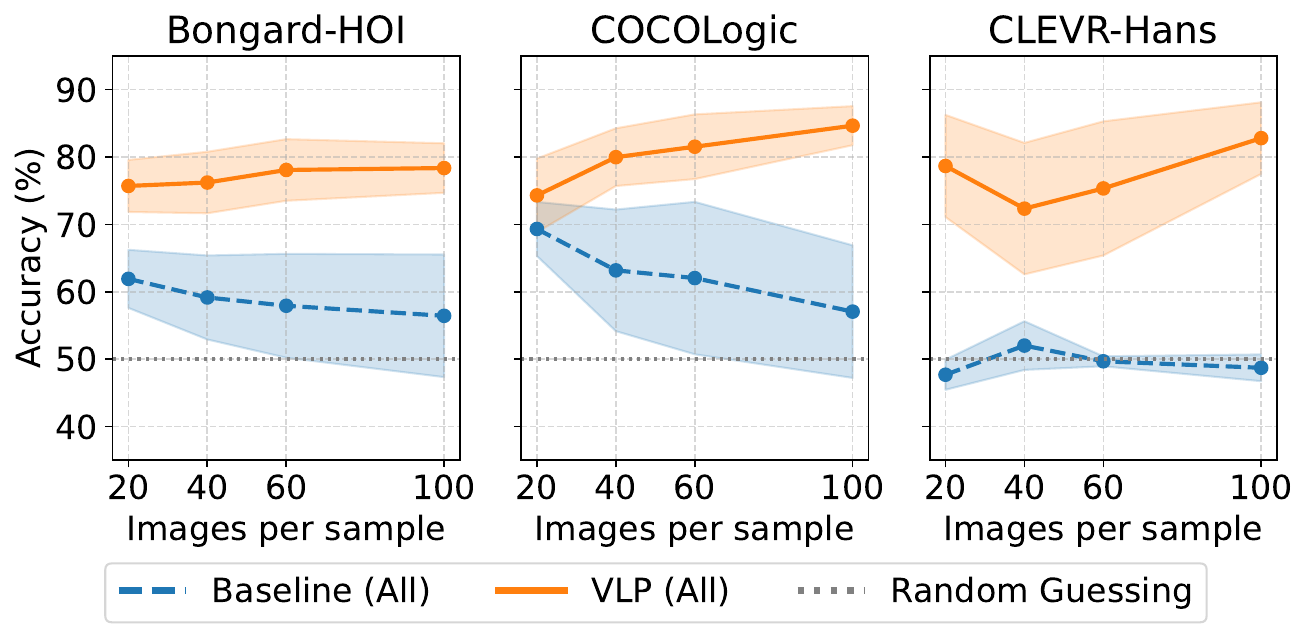}
    \caption{\textbf{VLP performance improves as more input images are provided}, in contrast to baselines, which stagnate or decline. Results are aggregated over models from Table 1.}
    \label{fig:more_imgs}
\end{figure}

\paragraph{VLPs reliably integrate larger numbers of samples (RQ3).} 
In many cases, having more samples for a specific task can provide valuable additional evidence for a concept and help refine it by excluding alternative explanations or shortcuts. In our third set of evaluations, we therefore examine how \method handles larger numbers of input images, particularly compared to the base VLM’s image processing capacity. As shown in \autoref{fig:more_imgs}, we increase the number of input images per task from 20 to 100 across the Bongard-HOI, COCOLogic, and CLEVR-Hans3 datasets. For a fair comparison, the number of test images remains fixed in all runs, and we only consider concepts with sufficient support samples (\cf~\autoref{app:data}). For each dataset, we report the average performance across all VLM models from \autoref{tab:vlp_results}, comparing the base models with and without \method.

We observe that performance improves for models using VLPs, as the additional evidence helps identify more accurate programs. For example, for the concept “\textit{carrying surfboard}” in Bongard-HOI (\cf Suppl.~\autoref{fig:example_surfing}), InternVL3-14B with 20 task examples discovers the program \symbols{((exists\_action (get\_actions IMG) holding))}, which does not specify the object being held but still covers 96\% of the few-shot examples. With 100 examples, however, the model retrieves a more precise program, including the actions \symbols{holding} and \symbols{walking} in combination with the object \symbols{surfboard} (\cf \autoref{app:qualitative_example_bongard_hoi}).


In contrast, the base VLMs do not benefit from an increased number of examples. Since they process all input images jointly, individual samples likely receive less attention, resulting in overly abstract learned representations. \Eg, for the previous \textit{surfing} task with 20 images, InternVL3-14B produces the concept “\textit{The activity involves walking with or actively surfing on a surfboard in a beach or ocean setting},” which is overly general and mistakenly includes negative examples (\textit{surfing} instead of \textit{carrying}).

Overall, these results highlight that \method can effectively leverage larger sets of input images by synthesizing more specific and semantically grounded programs, while base VLMs struggle to capitalize on the additional data. We hence answer RQ3 affirmatively.

\paragraph{VLPs enable interaction with the Solution Space of the Program (RQ4).}
An important feature of VLPs is their inherent modularity, which allows users to inspect and interact with individual processing steps. This enables targeted debugging of failure cases and refinement of solutions through step-specific feedback. To investigate how this facilitates intuitive interactions with the task's solution space, valuable for model debugging and shortcut mitigation \citep{stammer2021right, SchramowskiSTBH20, GeirhosJMZBBW20, steinmann2024navigating}, we investigate the solutions of the models InternVL3-8B and Qwen3-VL on the dataset CLEVR-Hans3 from \autoref{tab:vlp_results}.

\begin{figure}
    \centering
    \includegraphics[width=0.88\linewidth]{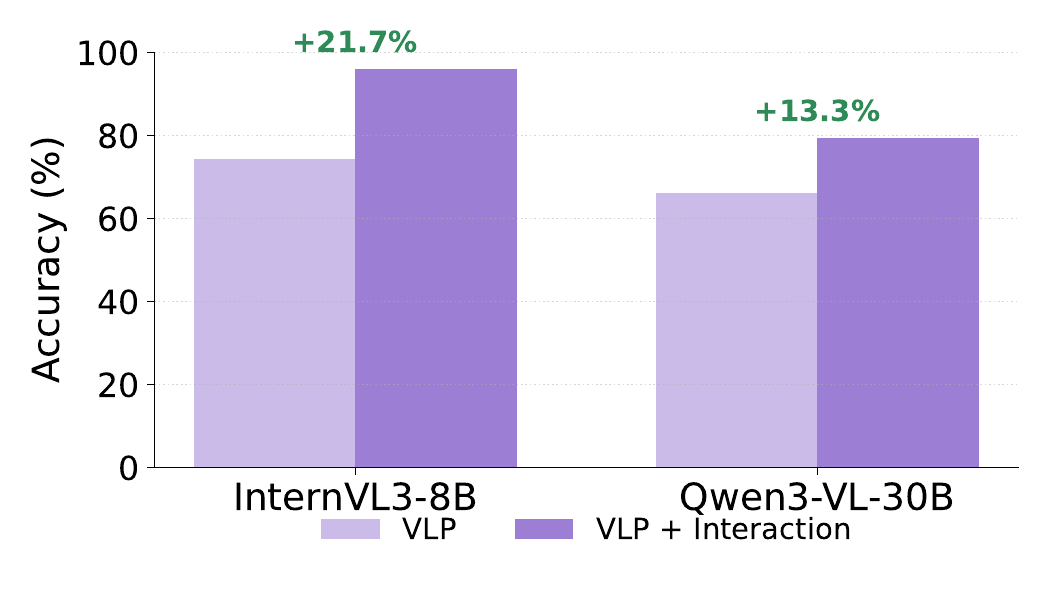}
    \caption{\textbf{VLP performance on CLEVR-Hans3 with DSL edits.} For InternVL3 \symbols{size}-related VLM functions were added, for Qwen3 shortcut-related colors (\symbols{red}, \symbols{gold}) removed.}
    \label{fig:interaction}
\end{figure}

Here, we observed that, \eg, InternVL3-8B seldom considers the size of the objects, which is a relevant property for the rules. Looking at the grounded properties and the object-property representations, we observed that even though the size properties, \textit{small} and \textit{large}, are discovered, they are not always used in the object-property representation obtained by \texttt{get\_objects}. An explanation for this could be that the concept \textit{size} is very relative and depends on the context. 
Fortunately, thanks to the flexibility of VLP one can very simply add VLM functions to the DSL that directly ask for the size of a given object (\cf \autoref{app:vlm_functions_size}) in \textit{relation} to the other objects. As \autoref{fig:interaction} (left) highlights, by incorporating these functions into the DSL the performance on the query samples improves substantially, \ie, InternVL3-14B can now utilize this bias to achieve 96\% accuracy.

In contrast, when inspecting the results of Qwen3-VL, we identified that the VLPs indeed make use of the VLM-proposed properties, however in some cases a VLP falsely considers the colors "red" and "gold" as relevant properties. Manual inspection of corresponding images (\eg of task \#2 in Suppl. \autoref{fig:pos_example_vlp_clevr}), indeed identified these features as potential shortcuts for the task. Equipped with such discoveries a human user can easily interact with the DSL by removing "red" and "gold" from the available properties and by this targeted revision improve the model's performance by $13.3\%$ (\cf \autoref{fig:interaction} (right)). 


These evaluations demonstrated that even with limited data, solution quality can be increased by incorporating additional task knowledge, \eg, in a human-in-the-loop setting. We therefore conclude RQ4 positively.

\section{Discussion}

Across our evaluations, we have observed that \method can be used to effectively improve visual inductive reasoning in terms of predictive performance improvements, but also failure analysis and model debugging. Below, we discuss additional considerations of our evaluations.

First, we investigated the impact of VLP’s occurrence-based symbol weighting in \autoref{app:distribution}. The results confirm its advantage over default uniform weighting, particularly in compositional settings.
Additionally, our comparative analysis in context of RQ1 revealed that InternVL3-8B w/ VLP achieved substantially stronger results than other configurations (\autoref{tab:vlp_results}), which we attribute to InternVL3-8B producing higher quality symbolic image representations, \cf~\autoref{app:failure_cases}). However, our failure analysis indicates that all models occasionally exhibited errors that limit VLP’s overall effectiveness (\cf~\autoref{app:failure_cases}). These include formatting errors (\eg, Kimi produced malformed symbolic representations in approximately 13\% of COCOLogic cases) and incomplete property descriptions, where VLMs omit task-relevant properties despite being capable of perceiving them (\eg, omitting "size" for CLEVR-Hans3, \cf \autoref{app:symbol_discovery}). Such perception failures constrain VLP performance by limiting available symbolic knowledge for reasoning. Notably, however, unlike end-to-end VLMs, where misclassifications remain opaque, VLPs enable tracing errors back to individual images within a task, providing actionable insights for model debugging and improvement.

Looking forward to improving perceptual grounding quality, VLMs could be prompted for each symbol individually rather than all at once (\eg, \symbols{get\_objects}), though this would substantially increase prompt overhead. We view the current approach as an effective trade-off, with future work potentially pre-selecting promising symbols during search to reduce computational costs. Additionally, \method currently lacks explicit object representations, limiting its effectiveness for object-centric concepts such as spatial relations. Extending \method with structured object representations would enable applications in complex expert domains like healthcare or mechanical engineering, where interactive human feedback during synthesis could prove particularly valuable. Future work could also investigate VLM robustness to out-of-distribution data and explore multi-model program synthesis, \eg, combining different VLMs or integrating classical algorithms within a unified DSL.

\section{Conclusion}
In this work, we introduced \methodfull (\method), a framework that integrates the perceptual flexibility of VLMs with the systematic reasoning capabilities of program synthesis. By leveraging VLMs to generate structured symbolic descriptions that can be compiled into executable programs, \method enable explicit composition, disambiguation, and, importantly, human-level interpretability beyond the natural language outputs of VLMs.
Our empirical results across synthetic and real-world datasets show that \method consistently improve generalization and predictive performance compared to direct prompting and structured prompting, even without the need for domain-specific encoders. Our work demonstrates that VLMs can be naturally integrated into a programmatic framework as callable functions rather than monolithic end-to-end predictors, bringing together the richness of neural vision-language representations with the reliability and controllabilty of symbolic program synthesis.



%% file: sec/acknowledgments.tex
\section*{Acknowledgments}
This work was supported by the Priority Program (SPP) 2422 in the subproject “Optimization of active surface design of high-speed progressive tools using machine and deep learning algorithms“ funded by the German Research Foundation (DFG). The Eindhoven University of Technology authors received support from their Department of Mathematics and Computer Science and the Eindhoven Artificial Intelligence Systems Institute.
Furthermore, this work has benefited from early stages of the Cluster of Excellence "Reasonable AI" funded by the German Research Foundation (DFG) under Germany’s Excellence Strategy (EXC-3057), funding will begin in 2026.

We gratefully acknowledge support from the hessian.AI Service Center (funded by the Federal Ministry of Research, Technology and Space, BMFTR, grant no. 16IS22091) and the hessian.AI Innovation Lab (funded by the Hessian Ministry for Digital Strategy and Innovation, grant no. S-DIW04/0013/003).

%% file: sec/X_suppl.tex
\begin{center}
\textbf{\Large Synthesizing Visual Concepts as Vision-Language Programs \\}
\vspace{0.5em}\text{\Large Supplementary Material} \\
\vspace{1.0em}

\end{center}
\setcounter{page}{1}

\noindent The following appendix provides supplementary materials referenced in the main text. We begin with an overview of the datasets used in the evaluations (\autoref{app:data}), followed by a discussion of qualitative examples (\autoref{app:qualitative_examples}). Additional supporting experiments are presented in \autoref{app:additional_experiments}, and failure cases are analyzed in \autoref{app:failure_cases}. We discuss limitations of the dataset Bongard-OpenWorld in \autoref{app:bongard-ow-dataset-issues}. Finally, we describe the DSL used for \method\ (\autoref{sec:dsl}) and list the prompts employed in the experiments (\autoref{app:prompts}).

\section{Datasets}\label{app:data}
In the following, we present the datasets used in the evaluations in more detail, along with representative samples.

\begin{figure*}[t!]
    \centering
    \includegraphics[width=0.8\linewidth]{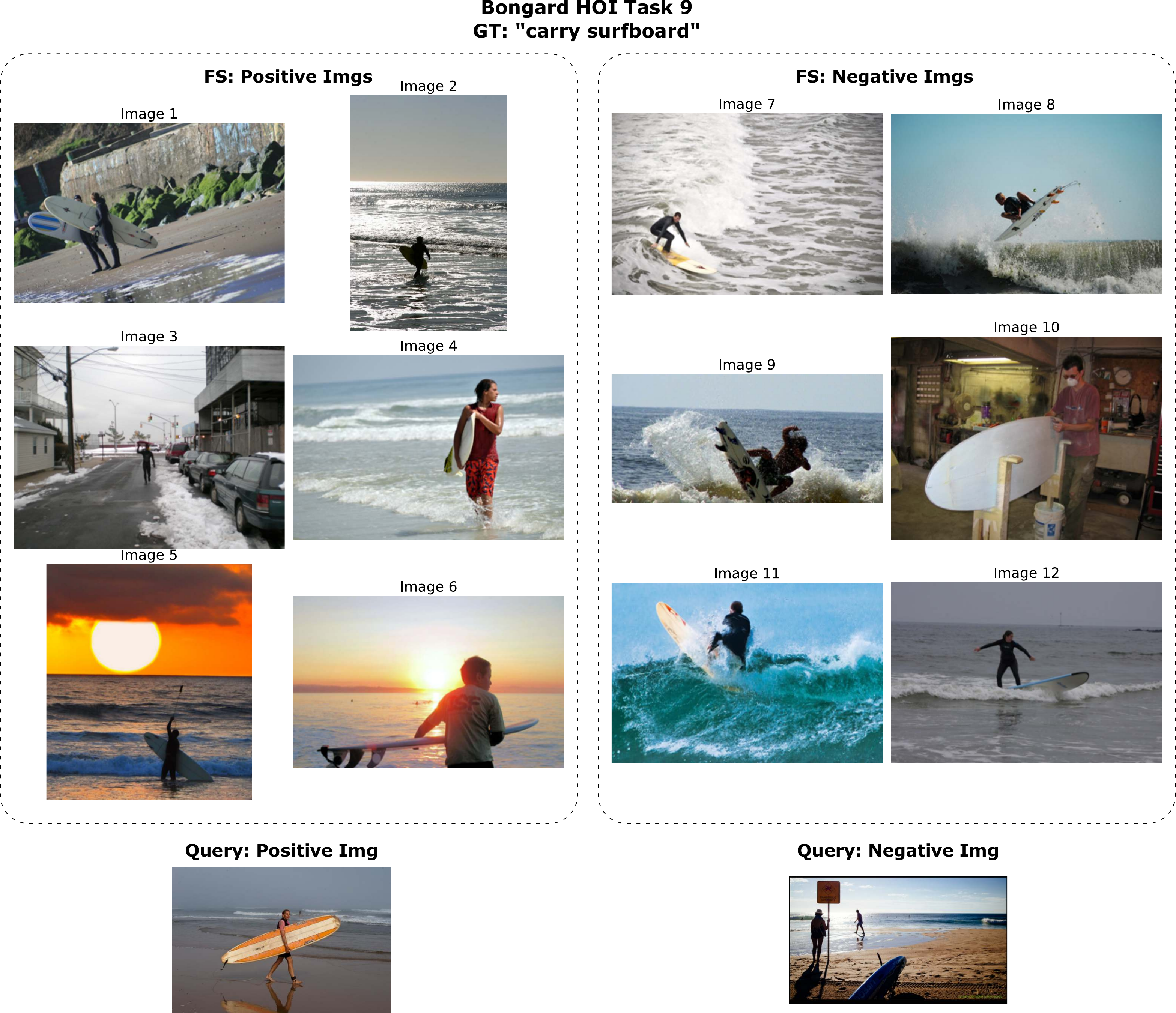}
    \caption{Example task from Bongard-HOI (Task 9, GT: ``carry surfboard''). Bongard-HOI focuses on human-object interactions in natural images, requiring models to distinguish specific actions or relationships between people and objects. Positive examples show people carrying surfboards, while negative examples depict other surfboard-related activities such as surfing or standing near surfboards without carrying them.}
    \label{fig:example_surfing}
\end{figure*}

\paragraph{Bongard-HOI.}
The Bongard-HOI dataset \citep{jiang2022bongard} is one of the real-world image datasets used for evaluating our method, specifically focusing on logical rules related to Human-Object Interactions (HOI). This dataset requires the model to induce rules based on how \textit{people are interacting with objects} in an image. Like all Bongard datasets used, each task provides $12$ samples from which the target rule should be induced: $6$ positive examples that conform to the latent rule (the rule set) and $6$ negative examples that violate the rule (the anti-rule set). 
We evaluate on all 166 test concepts of the four different test splits. For \autoref{tab:vlp_results}, we take the first sample of each rule. For RQ3 we reduce the test concepts to $67$, only keeping those concepts that have enough few-shot examples ($50$ or more for positive and negative set respectively). To achieve this, we collect all available images per rule and discard the problems that have fewer than 100 unique few-shot examples. The number of test samples is set to 4. We ensure that test examples are not present in the few-shot examples. This reduces the dataset from $166$ to $67$ problems. An example concept that is tested in RQ1 and RQ3 both is depicted in \autoref{fig:example_surfing}.

\begin{figure*}[t!]
    \centering
    \includegraphics[width=0.8\linewidth]{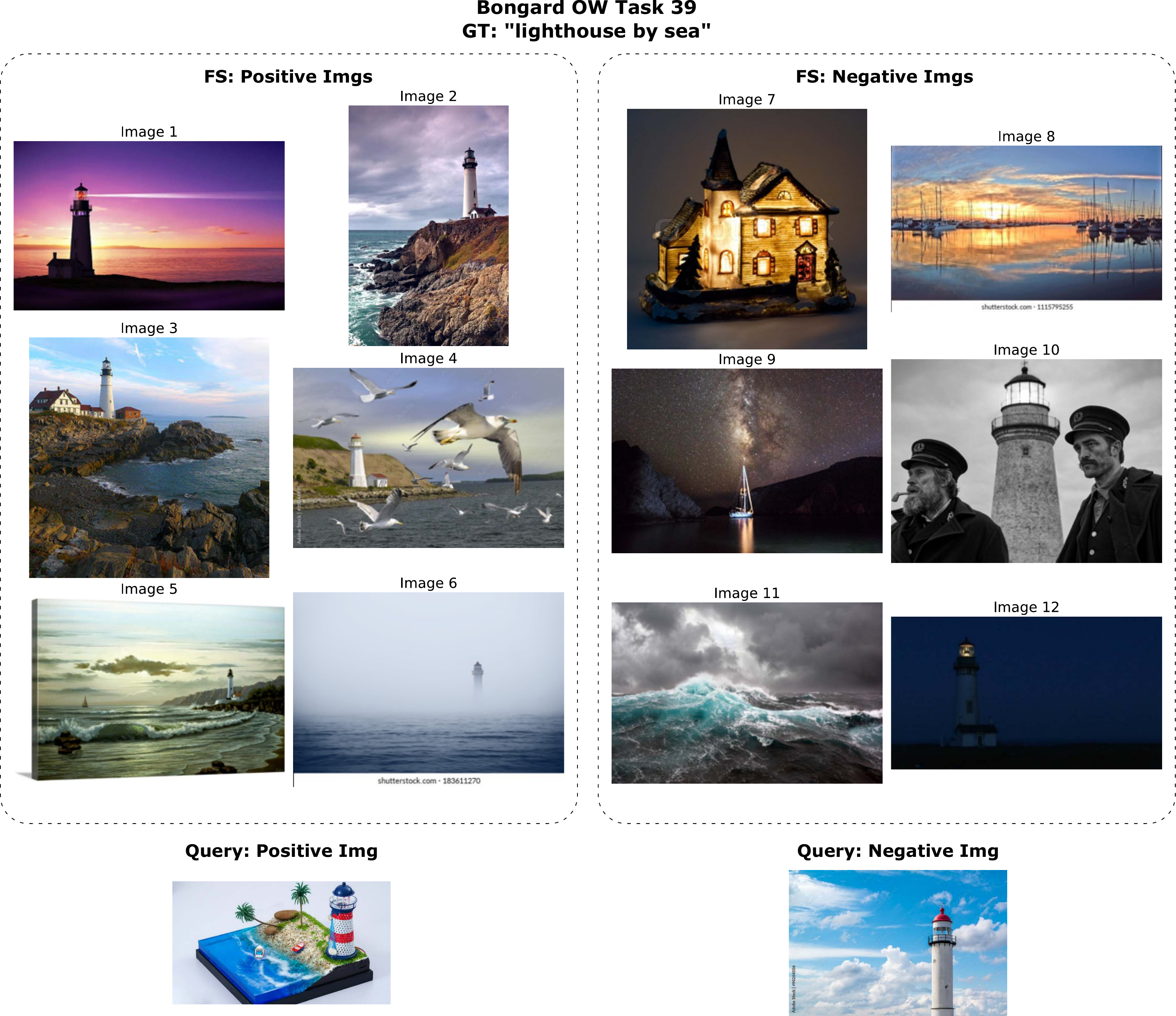}
    \caption{Example task from Bongard-OpenWorld (Task 39, GT: ``lighthouse by sea''). Each task consists of 6 positive few-shot images (left) demonstrating the target concept, 6 negative few-shot images (right) showing contrasting examples, and two query images (bottom) to be classified. Models must induce the visual rule from the few-shot examples and apply it to novel queries.}
    \label{fig:bongard-ow-good}
\end{figure*}

\paragraph{Bongard-OpenWorld.} The Bongard-OpenWorld dataset \citep{wubongard} is based on real-world images and concepts, designed to extend the challenge of rule induction to broader, open-world concepts and complex relational patterns. It shares the fundamental Bongard problem structure, where the goal is to induce a hidden logical rule from a small, balanced set of visual examples. Each task consists of $12$ samples in total: $6$ positive examples conforming to the latent rule, and $6$ negative examples that violate it. An example is depicted in \autoref{fig:bongard-ow-good}.
We evaluate our method on all $200$ test samples of the dataset. 

\begin{figure*}[t!]
    \centering
    \includegraphics[width=0.8\linewidth]{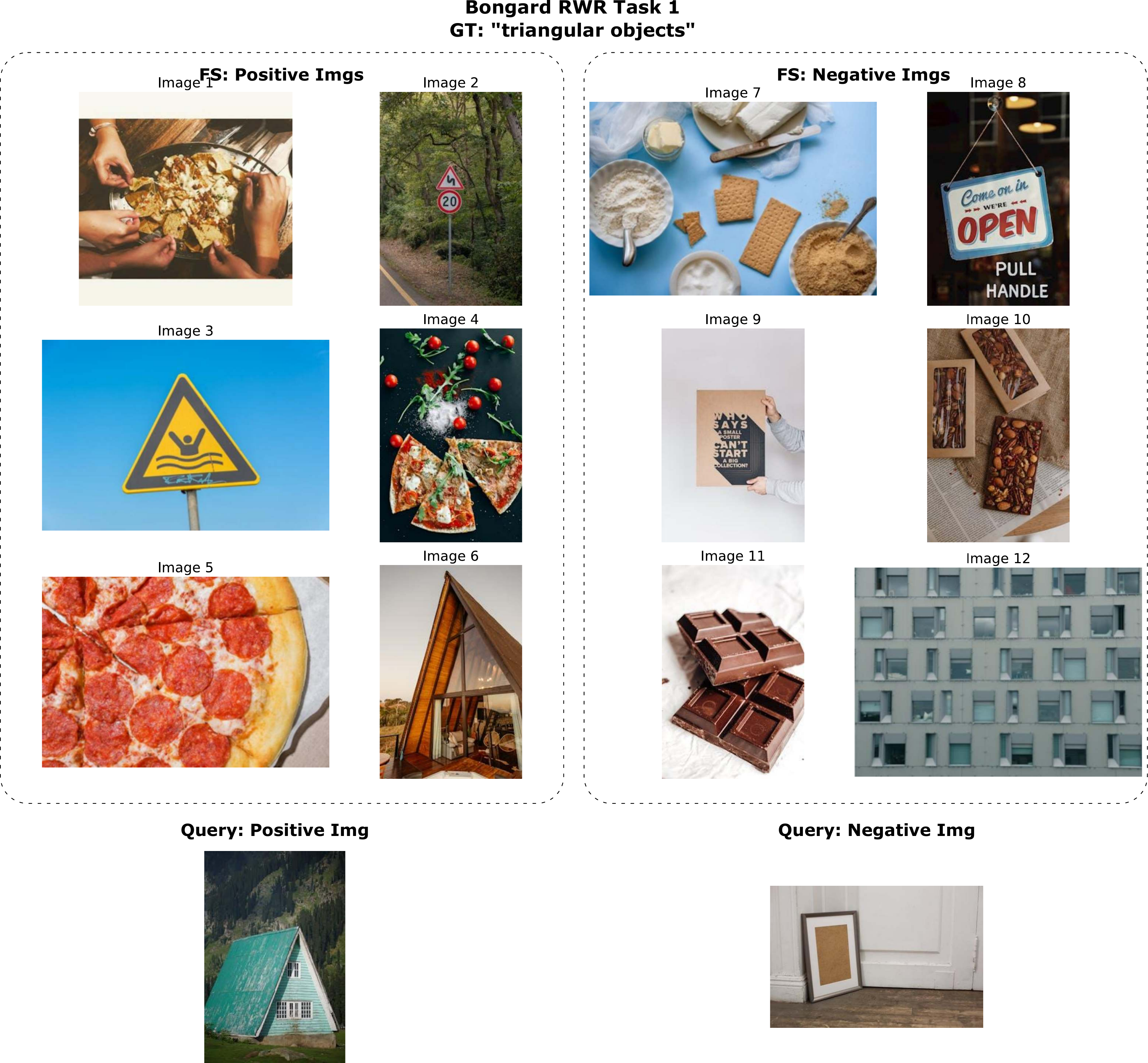}
    \caption{Example task from Bongard-RWR (Task 1, GT: ``triangular objects''). Similar to Bongard-OW, each task provides 6 positive and 6 negative few-shot examples, followed by two query images to classify. Bongard-RWR features real-world photographic images with more complex visual scenes and contextual variation compared to simplified geometric tasks.}
    \label{fig:rwr-good-example}
\end{figure*}

\begin{figure*}[t!]
    \centering
    \includegraphics[width=0.8\linewidth]{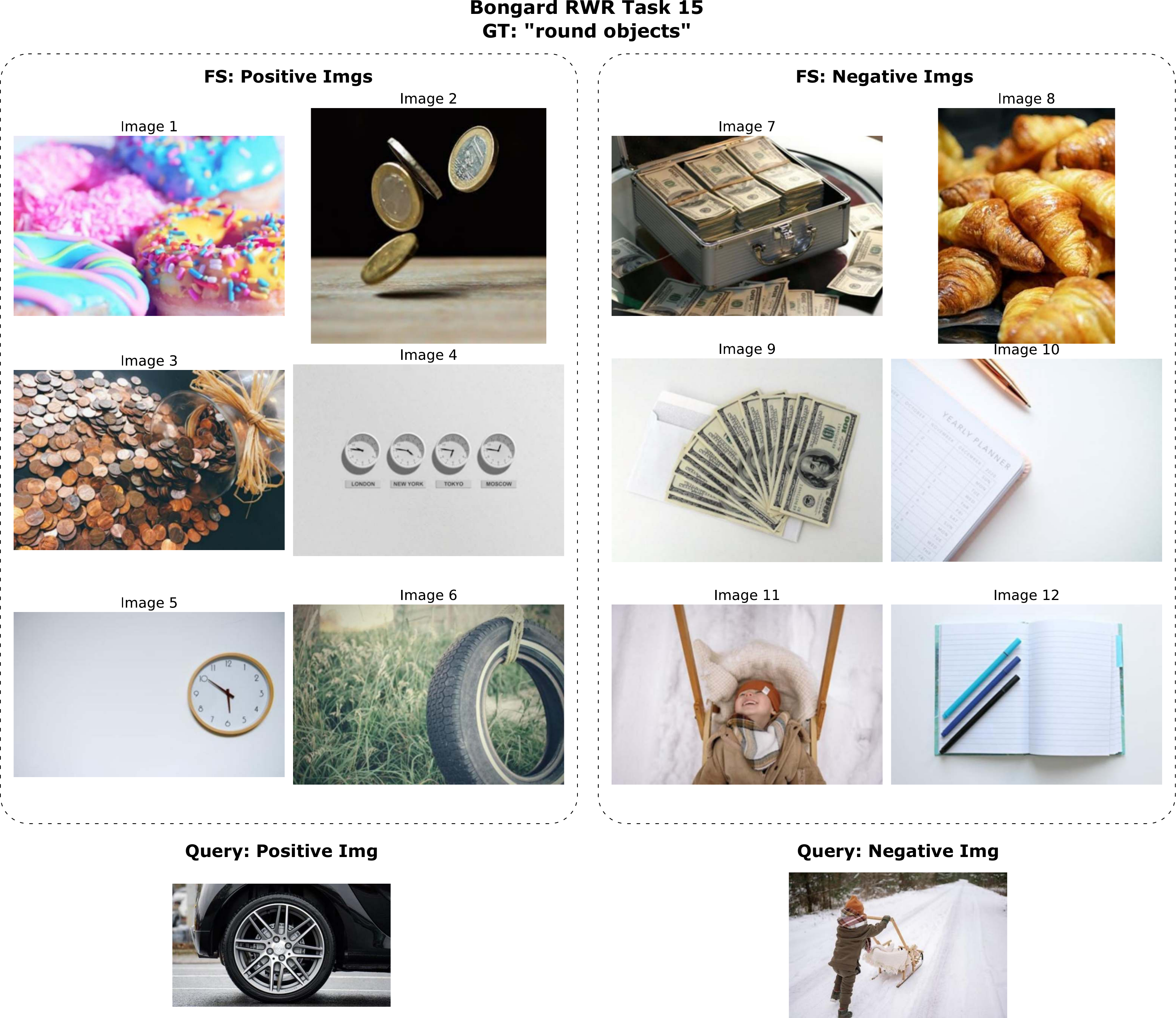}
    \caption{Bongard-RWR Task 15 (GT: ``round objects''). Positive examples feature prominent circular or round objects (donuts, coins, clocks, wheels), while negative examples show rectangular or non-round items (money cases, croissants, notebooks). The task requires identifying roundness as the distinguishing visual property across diverse object categories and contexts.}
    \label{fig:rwr-good-example-2}
\end{figure*}

\paragraph{Bongard-RWR} The Bongard-RWR dataset \citep{Pawlonka25bongardrwr} is also based on real-world images, specifically focusing on abstract visual concepts derived from the original Bongard problems. It is utilized to test the induction of abstract and complex rules within natural, diverse imagery. Following the standard format, each task instance provides $12$ samples ($6$ positives and $6$ negatives) from which the underlying logical rule must be determined.
We evaluate on all $60$ concepts of the dataset and take the first sample per concept (same as Bongard-HOI). Examples of the dataset are provided in \autoref{fig:rwr-good-example} and \autoref{fig:rwr-good-example-2}.

\paragraph{COCOLogic.}
The COCOLogic dataset \citep{steinmann2025object} is selected for its inclusion of complex logical rules applied to real-world images. Built upon the widely used COCO dataset, it introduces logic-based classification tasks that require reasoning over object co-presence and object counts. 
We construct ten distinct tasks from COCOLogic, one for each dataset class. Positive samples are drawn from the target class to ensure all relevant objects appear at least once in the few-shot examples, while negative samples are taken from the remaining classes. Each few-shot task includes $20$ balanced samples. For testing, we select up to ten query samples per task for both positive and negative classes, although some classes contain fewer available examples.
For RQ3, similar to Bongard-HOI, we use a subset of $8$ tasks to ensure a sufficient number of few-shot examples. Overall, COCOLogic provides a challenging benchmark for evaluating model performance on complex logical reasoning tasks within diverse, natural image contexts.

\begin{figure*}[t!]
    \centering
    \includegraphics[width=0.8\linewidth]{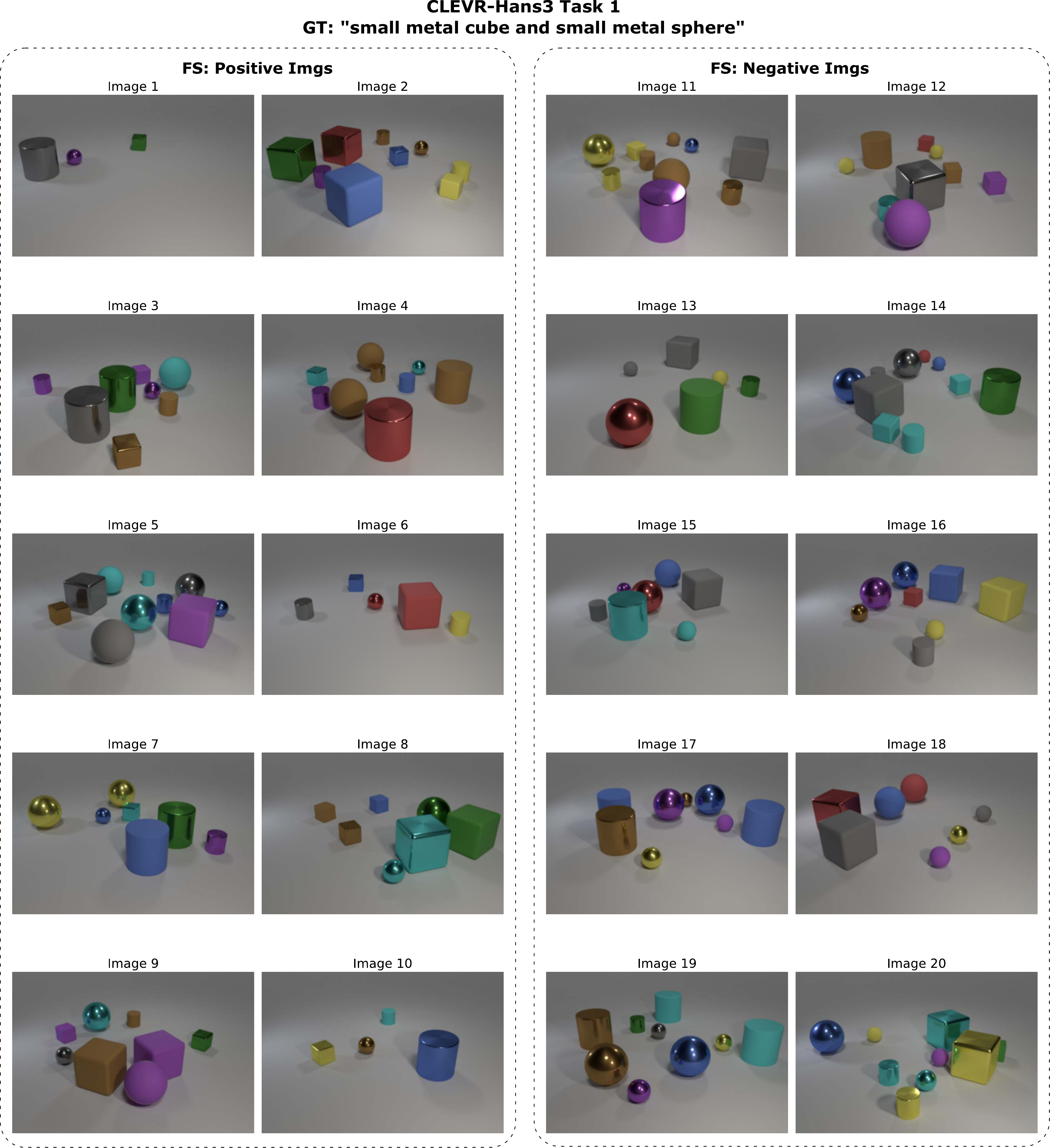}
    \caption{Example task from CLEVR-Hans3 (Task 1, GT: ``small metal cube and small metal sphere''). Each task contains 10 positive and 10 negative few-shot examples (shown), plus 10 positive and 10 negative query images (not shown for space). The dataset uses synthetically rendered 3D scenes with controlled object properties (size, material, shape, color), enabling systematic evaluation of compositional reasoning with multi-attribute conjunctive rules. A shortcut that is discovered by Qwen3-VL in these few-shot examples is the rule "\textit{red object XOR not gold metallic cylinder}" (true for all images except for image 6).}
    \label{fig:pos_example_vlp_clevr}
\end{figure*}

\paragraph{CLEVR-Hans3} For the synthetic dataset component, we utilize CLEVR-Hans3 \citep{stammer2021right}. This dataset is a variant of the CLEVR~\citep{johnson2017clevr} dataset, specifically constructed to enforce complex logical rules. We leverage the three primary classes present in CLEVR-Hans3 to construct three corresponding tasks, each requiring the induction of complex logical rules related to object color, shape, size, and material. For the support set, we take $20$ balanced samples per task, same for the query samples. 
We utilize the original confounded validation set of CLEVR-Hans3 in our evaluations as a test set, thus turning the originally confounded classification task into a standard one. An example is provided in \autoref{fig:pos_example_vlp_clevr}.

\begin{figure*}[t!]
    \centering
    \includegraphics[width=0.8\linewidth]{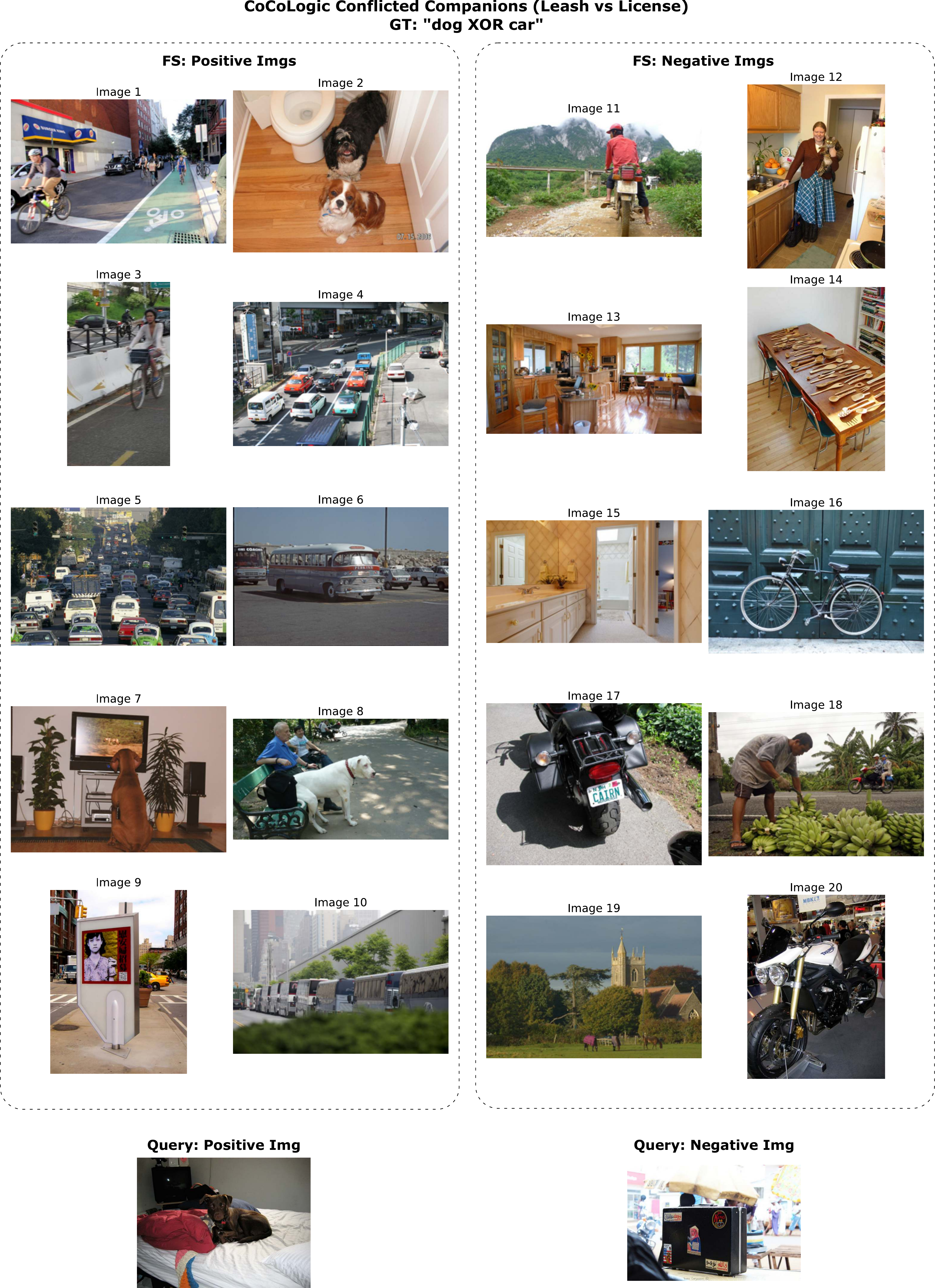}
    \caption{Example task from COCOLogic (``Conflicted Companions'', GT: ``dog XOR car''). Each task contains 10 positive and 10 negative few-shot examples (shown), plus 2 query images (shown at bottom). The dataset uses natural images from COCO to test logical reasoning with Boolean operators (AND, OR, XOR, NOT). This task requires identifying images containing either dogs or cars, but not both.}
    \label{fig:enter-label}
\end{figure*}

\section{Qualitative Examples}\label{app:qualitative_examples}

In the following we show an example of results from \method on a Bongard-HOI task (\autoref{app:qualitative_example_bongard_hoi}). 

\subsection{Bongard-HOI Example}\label{app:qualitative_example_bongard_hoi}

\method is not entirely free from shortcut learning. However, we consider this to be a limitation inherent to the nature of the underlying problems, such as the scarcity of hard negative examples and the presence of ambiguous concepts. We illustrate this with a qualitative example of shortcuts identified in \method below.

For the concept “carry surfboard” in Bongard-HOI, which was analyzed in RQ1 and RQ3, we observed that with only 12 and 20 support samples, our method retrieved the following program in one run with InternVL-14B:

\[
\symbols{(exists\_action(get\_actions IMG) holding)}
\]
This program checks if the positive examples all include the action "holding" and correctly classifies all few-shot examples but fails to specify the object being held.
InternVL3-8B, discovers a composition of two actions, holding or walking:

\[ 
\begin{split}
\symbols{(or (exists\_action (get\_actions IMG) holding)} \\ \symbols{(exists\_action (get\_actions IMG) walking))} 
\end{split}
\]

In the RQ3 experiments, we increase the number of few-shot examples, thereby providing stronger evidence for the target concept. Consequently, the retrieved programs become more refined. For example, with 100 few-shot examples, InternVL3-14B retrieves:

\[
\begin{split}
\symbols{(or (exists\_action\_with\_object (get\_actions var0) holding surfboard)} \\
\symbols{(exists\_action\_with\_object (get\_actions var0) standing surfboard))}
\end{split}
\]
checking, if there is either the action \textit{holding surfboard} or \textit{standing} with \textit{surfboard} present. InternVL3-8B retrieves the program:
\[
\begin{split}
\symbols{(or (exists\_action\_with\_object (get\_actions IMG) walking surfboard)} \\
\symbols{(exists\_action\_with\_object (get\_actions IMG) holding surfboard))}.
\end{split}
\]
which checks if there is the action \textit{walking} with \textit{surfboard} or \textit{holding surfboard}.

\clearpage
\section{Additional Experiments}\label{app:additional_experiments}

To complement the results presented in the main paper, this appendix provides a set of additional experiments designed to deepen our understanding of how \method behaves under different design choices and ablations. These analyses address four key questions:
\begin{itemize}
    \item How much of \method's performance stems from its structured prompting strategy as opposed to direct end to end reasoning with VLMs? (\cf \autoref{app:reasoning_ablation})
    \item Does weighting symbols by their occurrence frequency lead to better programs than using a uniform distribution? (\cf \autoref{app:distribution})
    \item Does stochastic generation during the VLM extraction stage matter, or would greedy decoding suffice? (\cf \autoref{app:greedy_decoding})
\end{itemize}

Together, these studies allow us to isolate the contribution of individual components, quantify their effect on performance, and verify the robustness of \method across modeling and decoding settings. The following subsections report these results and discuss their implications.

\subsection{Ablation of Program Synthesis}\label{app:reasoning_ablation}
We examine whether structured symbolic representations of images provide an advantage over reasoning directly with the images. To achieve this, we introduce a second baseline in which VLMs receive the structured representations generated by the VLM functions of \method, rather than the raw images. This setup allows a more direct comparison of rule induction, since both approaches operate on the same underlying information. The results are presented in \autoref{tab:vlp_results_with_structure_baseline} as "w/ VLM functions" together with the original results from (RQ1). For Qwen3-VL we report one seed only, since the model got stuck in reasoning traces and was not able to formulate a final response in over 60\% of the cases.

Interestingly, this alternative increases performance slightly in some cases, but also performs worse than the standard baseline in others. On average, it is not able to come close to the performance of \method. Overall, the results indicate that rules induced by \method are by far more reliable than those obtained by prompting the model directly, even when structured symbolic representations are provided.
\begin{table*}[h]
    \centering
    \caption{\textbf{Comparison of base VLMs, base VLMs with VLM functions and VLMs with VLP (averaged over three runs).} Balanced accuracy (\%) on 6-shot Bongard tasks and 10-shot logical reasoning benchmarks. Best results per model are shown in \textbf{bold}; overall best results per dataset column are \underline{underlined}. *Qwen3-VL with one seed only, generation did not terminate.}
    \label{tab:vlp_results_with_structure_baseline}
    \vskip 0.4em
    \small
    \resizebox{0.98\linewidth}{!}{
    \begin{tabular}{lllllll}
    \toprule
    \textbf{Model} & \textbf{Avg.} &
    \multicolumn{3}{c}{\textbf{Bongard Tasks (6-shot)}} &
    \multicolumn{2}{c}{\textbf{Logical Reasoning (10-shot)}} \\
    \cmidrule(lr){3-5} \cmidrule(lr){6-7}
     &  & Bongard-HOI & Bongard-OW & Bongard-RWR & COCOLogic & CLEVR-Hans3 \\
    \midrule
    InternVL3-8B & $57.4$ & $60.5$ & $59.2$ & $47.2$ & $71.5$ & $48.3$ \\
    \quad w/ VLM functions & $56.6$ & $60.5$ & $56.2$ & $50.0$ & $70.2$ & $46.1$ \\
    \rowcolor{gray!10} \quad w/ \textbf{VLP} &
        \textbf{\underline{70.9}} {\footnotesize\textcolor{darkgreen}{(+13.5)}} &
        \textbf{\underline{77.7}} {\footnotesize\textcolor{darkgreen}{(+17.2)}} &
        \textbf{67.5} {\footnotesize\textcolor{darkgreen}{(+8.3)}} &
        \textbf{53.9} {\footnotesize\textcolor{darkgreen}{(+6.7)}} &
        \textbf{\underline{81.0}} {\footnotesize\textcolor{darkgreen}{(+9.5)}} &
        \textbf{74.4} {\footnotesize\textcolor{darkgreen}{(+26.1)}} \\
        \midrule
    InternVL3-14B & $61.5$ & $66.9$ & $62.5$ & $51.4$ & \textbf{78.4} & $48.3$ \\
    \quad w/ VLM functions & $61.5$ & $70.8$ & $62.5$ & $50.0$ & $74.0$ & $50.0$ \\
    \rowcolor{gray!10} \quad w/ \textbf{VLP} &
        \textbf{68.3} {\footnotesize\textcolor{darkgreen}{(+6.8)}} &
        \textbf{74.3} {\footnotesize\textcolor{darkgreen}{(+7.4)}} &
        \textbf{64.7} {\footnotesize\textcolor{darkgreen}{(+2.2)}} &
        \textbf{52.8} {\footnotesize\textcolor{darkgreen}{(+1.4)}} &
        $76.7$ {\footnotesize\textcolor{gray}{(-1.7)}} &
        \textbf{73.3} {\footnotesize\textcolor{darkgreen}{(+25.0)}} \\
        \midrule
    Kimi-VL-A3B-Instruct & $58.5$ & $59.8$ & $58.6$ & $46.4$ & \textbf{77.9} & $50.0$ \\
    \quad w/ VLM functions & $55.9$ & $57.3$ & $54.2$ & $53.3$ & $62.2$ & $52.2$ \\
    \rowcolor{gray!10} \quad w/ \textbf{VLP} &
        \textbf{65.5} {\footnotesize\textcolor{darkgreen}{(+7.0)}} &
        \textbf{69.4} {\footnotesize\textcolor{darkgreen}{(+9.6)}} &
        \textbf{59.4} {\footnotesize\textcolor{darkgreen}{(+0.8)}} &
        \textbf{52.5} {\footnotesize\textcolor{darkgreen}{(+6.1)}} &
        $70.1$ {\footnotesize\textcolor{gray}{(-7.8)}} &
        \textbf{76.1} {\footnotesize\textcolor{darkgreen}{(+26.1)}} \\
        \midrule
    Qwen2.5-VL-7B-Instruct & $60.1$ & $65.2$ & $\textbf{66.2}$ & \textbf{49.7} & $73.2$ & $46.1$ \\
    \quad w/ VLM functions & $56.9$ & $57.3$ & $55.5$ & $53.9$ & $61.9$ & $56.1$ \\
    \rowcolor{gray!10} \quad w/ \textbf{VLP} &
        \textbf{69.5} {\footnotesize\textcolor{darkgreen}{(+9.4)}} &
        \textbf{68.8} {\footnotesize\textcolor{darkgreen}{(+3.6)}} &
        $62.9$ {\footnotesize\textcolor{gray}{(-3.3)}} &
        $49.2$ {\footnotesize\textcolor{gray}{(-0.5)}} &
        \textbf{80.5} {\footnotesize\textcolor{darkgreen}{(+7.3)}} &
        \textbf{\underline{86.1}} {\footnotesize\textcolor{darkgreen}{(+40.0)}} \\
        \midrule
     Qwen3-VL-30B-A3B-Instruct & $63.4$  & $69.0$ & \textbf{\underline{68.5}} & $55.8$ & $73.9$ & $50.0$  \\
     \quad w/ VLM functions*  & $56.0$ & $60.2$ & $57.5$ & $50.8$ & $61.5$ & $50.0$ \\
    \rowcolor{gray!10} \quad w/ \textbf{VLP} &
        \textbf{68.9} {\footnotesize\textcolor{darkgreen}{(+5.5)}} &
        \textbf{74.5} {\footnotesize\textcolor{darkgreen}{(+5.5)}} &
        $66.3$ {\footnotesize\textcolor{gray}{(-2.2)}} &
        \textbf{\underline{58.3}} {\footnotesize\textcolor{darkgreen}{(+2.5)}} &
        \textbf{79.1} {\footnotesize\textcolor{darkgreen}{(+5.2)}} &
        $\textbf{66.1}$ {\footnotesize\textcolor{darkgreen}{(+16.1)}} \\
    \bottomrule
    \end{tabular}
    }
\end{table*}

\subsection{Uniform vs. Occurrence-based weighted distribution}\label{app:distribution}

In this section, we investigate whether our proposed occurrence-based symbol weighting improves the performance of the VLP compared to a uniform symbol distribution. \autoref{tab:distribution_comparison} reports the corresponding delta values. Overall, the weighted distribution leads to performance gains in most settings, indicating that emphasizing frequently observed symbols in the positive examples helps the model induce more accurate rules. 


\begin{table}[h]
    \centering
        \caption{Difference of using occurence-based weighting instead of uniform for symbols.}
    \label{tab:distribution_comparison}
        \resizebox{0.99\linewidth}{!}{
\begin{tabular}{lcccccc}
\toprule
         & Average & Bongard-HOI & Bongard-OW & Bongard-RWR & COCOLogic & CLEVR-Hans3 \\
\midrule
InternVL3-8B & $+1.1$ & $+0.5$ & $+1.0$ & $0.0$ & $+0.3$ & $+3.8$ \\
InternVL3-14B & $+0.2$ & $+1.3$ & $+0.4$ & $-0.8$ & $-0.3$ & $+0.5$ \\
Kimi-VL-A3B-Instruct & $+0.3$ & $+1.2$ & $+0.3$ & $+2.5$ & $+1.8$ & $-4.5$ \\
Qwen2.5-VL-7B-Instruct & $-0.9$ & $-0.5$ & $+0.7$ & $-3.3$ & $-1.0$ & $-0.6$ \\
Qwen3-VL-30B-A3B-Instruct & $+0.6$ & $+0.1$ & $0.0$ & $+0.2$ & $+2.1$ & $+0.5$ \\
\bottomrule
\end{tabular}
}

\end{table}

\subsection{Comparing Deterministic and Sampling Based Decoding}\label{app:greedy_decoding}

In \autoref{tab:vlp_no_sampling}, we re-evaluate the setups from \autoref{tab:vlp_results} using greedy decoding instead of sampling, both for the baseline model and for the \method prompts. The relative performance pattern across models remains largely unchanged, indicating that the improvements from \method do not depend on stochastic decoding but arise from the structure of the prompting itself.

Greedy decoding yields a small decline for the baseline with Intern-VL3 and Qwen2.5-VL models, while Kimi-VL and Qwen3-VL benefit slightly from it. Interestingly, while the results for \method with InternVL3-8B slightly decrease, for the other models, the results improve. Kimi-VL increases from an average of $65.5$ to $67.7$ under greedy decoding, and Qwen3-VL even gains an improvement from $68.9$ to $71.4$. These gains suggest that \method interacts well with deterministic generation, likely because the structured prompts restrict the model’s search space and reduce the chance of drifting into suboptimal continuations.

However, the previous best score of InternVL3-8B under sampling at $70.9$ reduces to $67.3$ in this setting. This indicates that, for some models, the small amount of exploration introduced by sampling might still be beneficial and allows the decoder to escape overly conservative predictions, \eg, during symbol grounding.

In general, greedy decoding produces more stable outputs when applying VLM functions, while sampling can provide useful diversity during symbol grounding by generating a broader set of candidate symbols. A balanced combination of both approaches may therefore offer the strongest performance, and exploring such hybrid strategies is an interesting direction for future work.

\begin{table*}[h]
    \centering
    \caption{\textbf{Comparison of baseline and VLP prompting with greedy decoding.} Accuracy (\%) across Bongard benchmarks and logical reasoning tasks. Improvements (\textcolor{darkgreen}{green} / \textcolor{gray}{gray}) denote changes relative to the baseline. Best results per model are shown in \textbf{bold}; overall best results per dataset column are \underline{underlined}.}
    \label{tab:vlp_no_sampling}
    \vskip 0.4em
    \small
    \resizebox{\linewidth}{!}{
    \begin{tabular}{lllllll}
    \toprule
    \textbf{Model} & \textbf{Avg.} &
    \multicolumn{3}{c}{\textbf{Bongard Tasks}} &
    \multicolumn{2}{c}{\textbf{Logical Reasoning}} \\
    \cmidrule(lr){3-5} \cmidrule(lr){6-7}
     &  & Bongard-HOI & Bongard-OP & Bongard-RWR & COCOLogic & CLEVR-Hans3 \\
    \midrule

    InternVL3-8B & $56.7$ & $58.4$ & $56.2$ & $\textbf{47.5}$ & $74.6$ & $46.7$ \\
    \rowcolor{gray!10} \quad w/ \textbf{VLP} &
    \textbf{67.3} {\footnotesize\textcolor{darkgreen}{(+10.6)}} &
    \textbf{\underline{79.2}} {\footnotesize\textcolor{darkgreen}{(+20.8)}} &
    \textbf{67.8} {\footnotesize\textcolor{darkgreen}{(+11.6)}} &
    42.5 {\footnotesize\textcolor{gray}{(-5.0)}} &
    \textbf{78.6} {\footnotesize\textcolor{darkgreen}{(+4.0)}} &
    \textbf{68.3} {\footnotesize\textcolor{darkgreen}{(+21.6)}} \\
    \midrule

    InternVL3-14B & $59.6$ & $67.2$ & $63.5$ & $48.3$ & $70.5$ & $48.3$ \\
    \rowcolor{gray!10} \quad w/ \textbf{VLP} &
    \textbf{69.1} {\footnotesize\textcolor{darkgreen}{(+9.5)}} &
    \textbf{75.0} {\footnotesize\textcolor{darkgreen}{(+7.8)}} &
    \textbf{66.2} {\footnotesize\textcolor{darkgreen}{(+2.7)}} &
    \textbf{51.7} {\footnotesize\textcolor{darkgreen}{(+3.4)}} &
    \textbf{\underline{81.1}} {\footnotesize\textcolor{darkgreen}{(+10.6)}} &
    \textbf{71.7} {\footnotesize\textcolor{darkgreen}{(+23.4)}} \\
    \midrule

    Kimi-VL-A3B-Instruct &  $59.2$ & $59.3$ & $\textbf{58.5}$ & $50.0$ & \textbf{78.4} & $50.0$ \\
    \rowcolor{gray!10} \quad w/ \textbf{VLP} &
    \textbf{67.2} {\footnotesize\textcolor{darkgreen}{(+8.0)}} &
    \textbf{70.2} {\footnotesize\textcolor{darkgreen}{(+10.9)}} &
    58.0 {\footnotesize\textcolor{gray}{(-0.5)}} &
    \textbf{53.3} {\footnotesize\textcolor{darkgreen}{(+3.3)}} &
    $72.8$ {\footnotesize\textcolor{gray}{(-5.6)}} &
    \textbf{81.7} {\footnotesize\textcolor{darkgreen}{(+31.7)}} \\
    \midrule

    Qwen2.5-VL-7B-Instruct & $60.0$ & $63.3$ & $\textbf{65.0}$ & $\textbf{51.7}$ & $75.0$ & $45.0$ \\
    \rowcolor{gray!10} \quad w/ \textbf{VLP} &
    \textbf{69.6} {\footnotesize\textcolor{darkgreen}{(+9.6)}} &
    \textbf{67.2} {\footnotesize\textcolor{darkgreen}{(+3.9)}} &
    63.5 {\footnotesize\textcolor{gray}{(-1.5)}} &
    48.3 {\footnotesize\textcolor{gray}{(-3.4)}} &
    \textbf{80.7} {\footnotesize\textcolor{darkgreen}{(+5.7)}} &
    \textbf{\underline{88.3}} {\footnotesize\textcolor{darkgreen}{(+43.3)}} \\
    \midrule

    Qwen3-VL-30B-A3B-Instruct & $63.7$ & $72.0$ & \textbf{\underline{69.5}} & $50.0$ & $76.8$ & $50.0$ \\
    \rowcolor{gray!10} \quad w/ \textbf{VLP} &
    \textbf{\underline{71.4}} {\footnotesize\textcolor{darkgreen}{(+7.7)}} &
    \textbf{74.4} {\footnotesize\textcolor{darkgreen}{(+2.4)}} &
    68.5 {\footnotesize\textcolor{gray}{(-1.0)}} &
    \textbf{\underline{60.0}} {\footnotesize\textcolor{darkgreen}{(+10.0)}} &
    \textbf{80.8} {\footnotesize\textcolor{darkgreen}{(+4.0)}} &
    \textbf{73.3} {\footnotesize\textcolor{darkgreen}{(+23.3)}} \\

    \bottomrule
    \end{tabular}
    }
\end{table*}



\section{Failure Cases}\label{app:failure_cases}

In this section we discuss potential failure cases of our method. These failures occur primarily for two reasons: (1) the VLM produces incorrect symbolic representations or misinterprets visual content (\autoref{app:generation_failures}), or (2) the VLM fails to retrieve the relevant image elements during symbol grounding (\autoref{app:symbol_discovery}).

\subsection{VLM Generation Quality}\label{app:generation_failures}
We evaluate the syntactic validity of the symbolic representations produced by the VLMs for \symbols{get\_objects} and \symbols{get\_actions} in \autoref{tab:object_extraction_results}. Models such as InternVL3-14B reliably generate parsable outputs, whereas others, like Kimi, produce numerous non-parsable representations, particularly on COCOLogic. These failures typically arise when the model enters a repetitive loop, repeating list elements without closing the list. In such cases, we automatically repair the list, though the resulting content is often less informative and may contain hallucinations. This behavior also helps explain why Kimi is the model that performs worst when paired with \method.

\begin{table}[h]
\centering
\caption{\textbf{Object} and \textbf{Action} Parse Rates Across Models and Datasets}
\label{tab:object_extraction_results}

\resizebox{0.7\linewidth}{!}{
\begin{tabular}{llrr}
\toprule
Dataset & Model & Object parse rate & Action parse rate \\
\midrule

\multirow{3}{*}{bongard-hoi}
& InternVL3-14B          & 1.00 & 1.00 \\
& InternVL3-8B           & 1.00 & 1.00 \\
& Kimi-VL-A3B-Instruct  & 0.96 & 0.95  \\
& Qwen2.5-VL-7B-Instruct & 0.98 & 0.99 \\
& Qwen3-VL-30B-A3B-Instruct & 1.00 & 1.00 \\

\midrule
\multirow{4}{*}{bongard-op}
& InternVL3-14B          & 1.00 & 1.00 \\
& InternVL3-8B           & 1.00 & 1.00 \\
& Kimi-VL-A3B-Instruct  & 0.94 & 0.93 \\
& Qwen2.5-VL-7B-Instruct & 0.98 & 0.98 \\
& Qwen3-VL-30B-A3B-Instruct & 0.99 & 1.00 \\

\midrule
\multirow{4}{*}{bongard-rwr}
& InternVL3-14B          & 1.00 & 1.00 \\
& InternVL3-8B           & 1.00 & 1.00 \\
& Kimi-VL-A3B-Instruct   & 0.97 & 0.73 \\
& Qwen2.5-VL-7B-Instruct & 0.98 & 1.00 \\
& Qwen3-VL-30B-A3B-Instruct & 0.99 & 1.00 \\

\midrule
\multirow{4}{*}{cocologic}
& InternVL3-14B          & 1.00 & 1.00 \\
& InternVL3-8B           & 1.00 & 1.00 \\
& Kimi-VL-A3B-Instruct   & 0.87 & 0.87 \\
& Qwen2.5-VL-7B-Instruct & 0.94 & 0.97 \\
& Qwen3-VL-30B-A3B-Instruct & 0.97 & 1.00 \\

\midrule
\multirow{4}{*}{CLEVR-Hans3}
& InternVL3-14B          & 1.00 & - \\
& InternVL3-8B           & 0.99 & - \\
& Kimi-VL-A3B-Instruct   & 0.99 & - \\
& Qwen2.5-VL-7B-Instruct & 1.00 & - \\
& Qwen3-VL-30B-A3B-Instruct & 1.00 & - \\

\bottomrule
\end{tabular}
}
\end{table}

\subsection{Symbol Discovery Quality}\label{app:symbol_discovery}

\method can only reason about symbols that have been retrieved during the \textit{symbol grounding} stage. Therefore, one obvious failure case of \method is that one or more relevant symbols for the visual concept are not retrieved in this initial step. To analyze how often this occurs, we conduct an experiment, checking the quality of the grounded symbols for each model and dataset. Hereby, we proceed as follows. If a dataset has objects, properties, or actions explicitly specified in the ground truth rule, such as Bongard-HOI, COCOLogic, and CLEVR-Hans, we leverage them to check the grounded symbols against the elements present in the rule. For the datasets Bongard-OpenWorld and Bongard-RWR, the visual concepts are more vague and not always bound to objects, \eg "\textit{living room}" or "\textit{aerial view}". Here, we ask an LLM to extract objects, properties, and actions from the rule to serve as a comparison to the grounded symbols of VLP, as an approximation.

Since there might be multiple terms with the same or close semantic meaning, we do not perform exact equality checking on the retrieved symbols but rather ask an LLM to judge whether the ground truth symbol is present in the symbols retrieved by \method. For both LLM usages described, we take the model "gpt-4o-2024-08-06".

\begin{figure}[h]
    \centering
    \begin{minipage}[b]{\textwidth}
        \centering
        \includegraphics[width=0.75\textwidth]{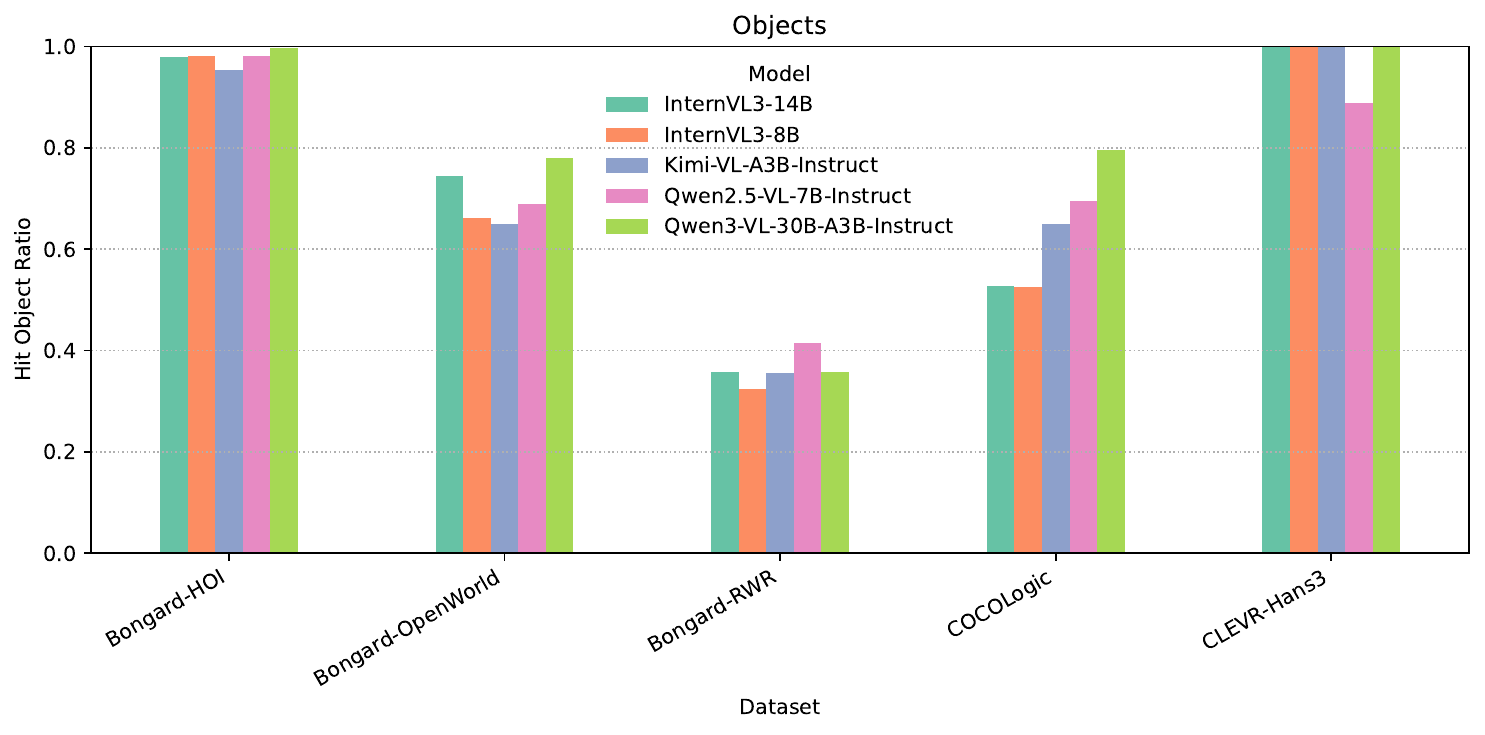}
        \label{fig:object-quality}
    \end{minipage}
    
    \begin{minipage}[b]{0.49\textwidth}
        \centering
        \includegraphics[width=\textwidth]{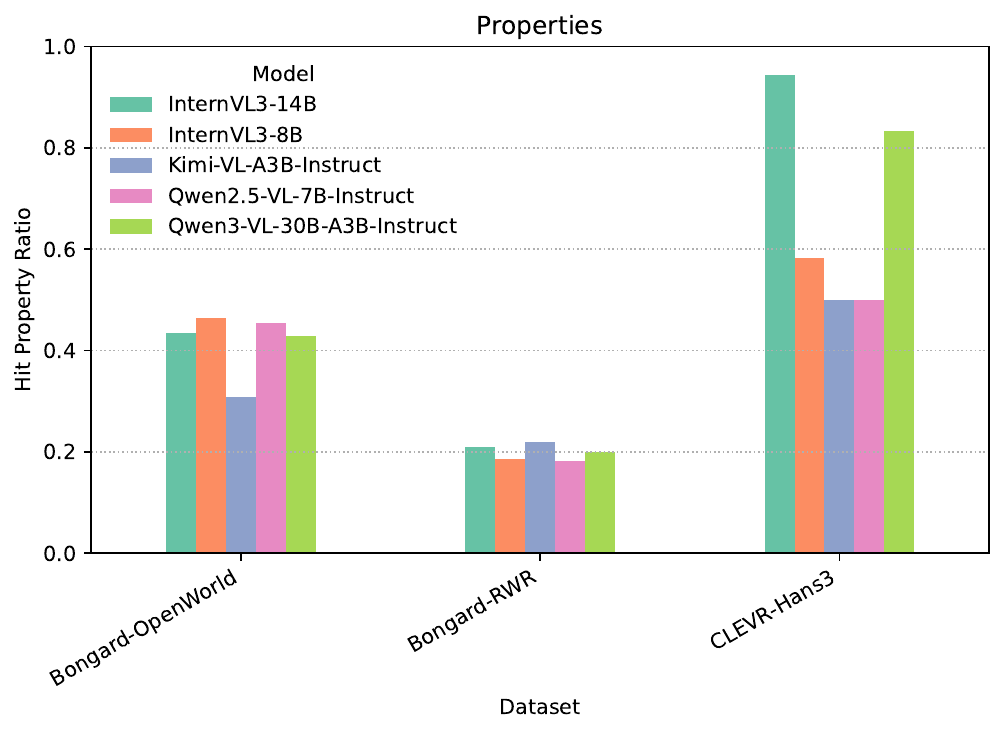}
        \label{fig:property-quality}
    \end{minipage}
    \hfill
    \begin{minipage}[b]{0.49\textwidth}
        \centering
        \includegraphics[width=\textwidth]{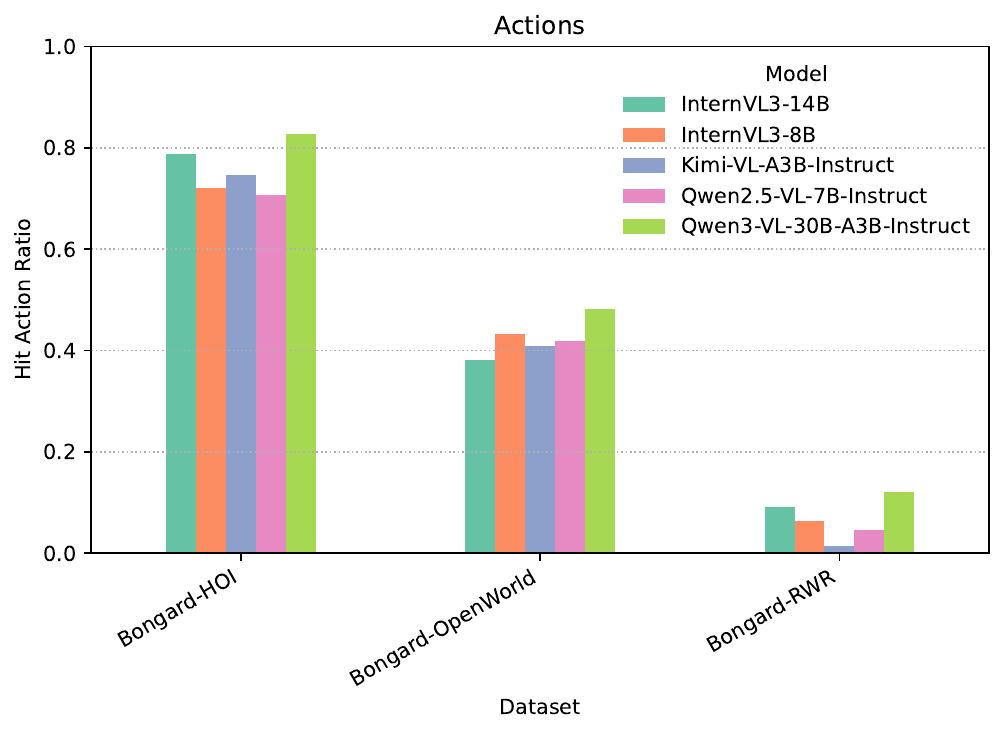}
        \label{fig:action-quality}
    \end{minipage}
    \caption{Hit ratios of the discovered object, property, and action symbols across all datasets. These values measure how often the symbols extracted by the VLM match the ground truth concepts. Datasets on which \method performs well in rule induction also show higher symbol quality in this analysis.}
    \label{fig:symbol-quality}
\end{figure}

The results of this analysis are displayed in \autoref{fig:symbol-quality} for objects, properties, and actions, respectively. Datasets for a symbol type are only considered if the symbol type is part of the ground truth concept. 

We observe that datasets on which \method achieves the strongest results also tend to exhibit higher quality ratios for the discovered symbols. This relationship is expected, as strong downstream performance implies that the model can reliably detect the visual aspects relevant to the ground-truth concept. Conversely, low symbol quality is likely a contributing factor to weak performance in the rule induction task. For example, Bongard-RWR shows the lowest quality ratios across all components, which aligns with its status as the most challenging dataset in our main evaluation.

While the ground truth symbols in this analysis are approximations generated by an LLM from the official Bongard solutions\footnote{\url{https://www.foundalis.com/res/bps/bongard_problems_solutions.htm}}, the overall trend remains plausible. The original Bongard concepts are well known to be difficult for modern VLMs, even when translated into real-world imagery \cite{wustbongard, Pawlonka25bongardrwr}.

\section{Dataset Quality Issues in Bongard-OpenWorld}\label{app:bongard-ow-dataset-issues}

During our experiments, we identified systematic annotation errors in the Bongard-OpenWorld dataset that fundamentally compromise both learning and evaluation. These issues fall into three categories: (1) mislabeled few-shot examples that contradict the ground-truth rule, (2) query images that violate the stated concept, and (3) inconsistent application of conjunctive rules. We document representative examples below to illustrate how these errors prevent reliable performance assessment.

\subsection{Inconsistent Few-Shot Labeling}

\begin{figure}[h]
    \centering
    \includegraphics[width=0.9\linewidth]{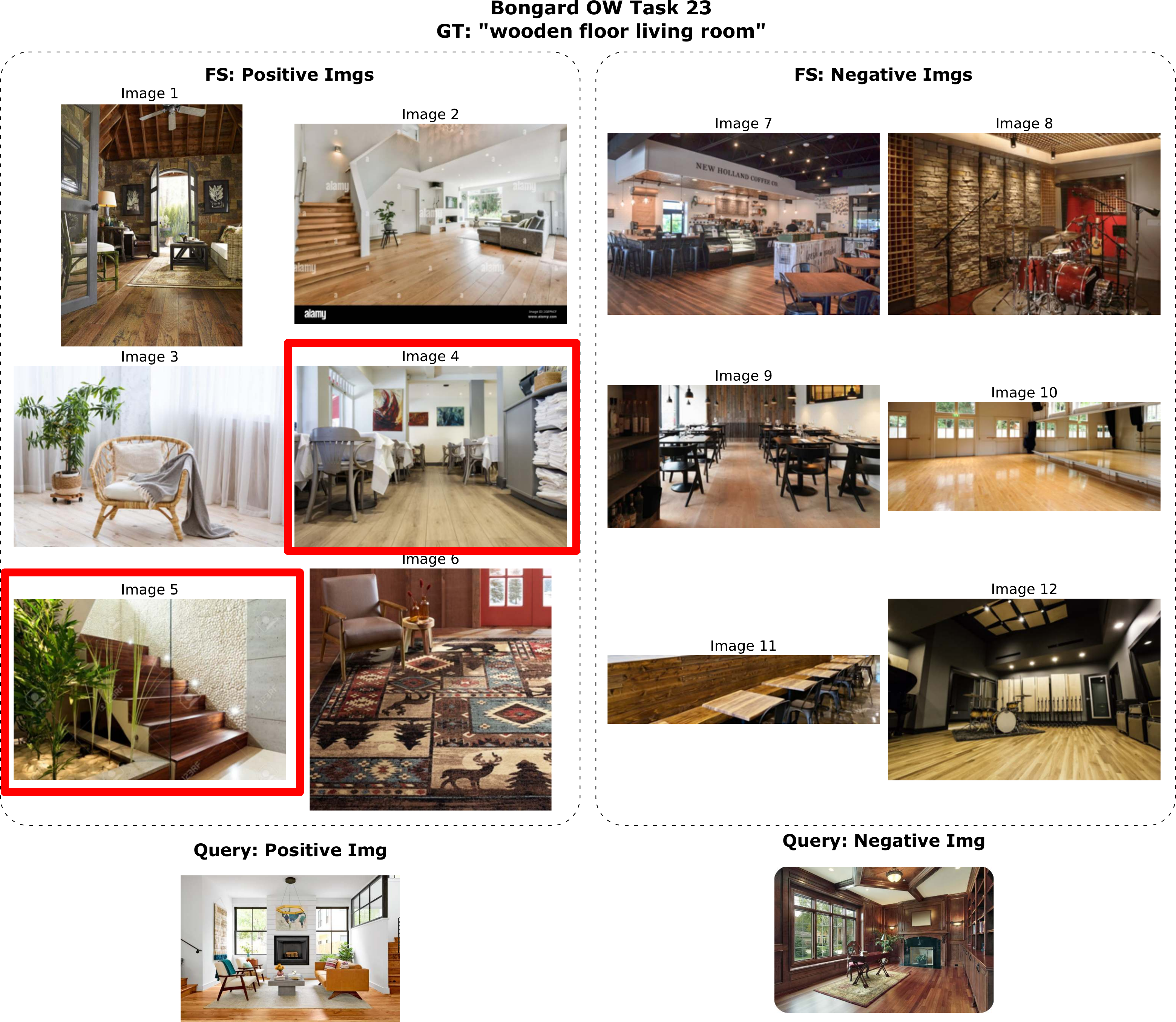}
    \caption{Annotation errors in Bongard-OW Task 23 reveal inconsistent rule application (GT: ``wooden floor living room''). Positive few-shot examples include a dining space (Image 4) and a staircase (Image 5, red boxes), which contain wooden floors but are not living rooms.}
    \label{fig:bongard-ow-23}
\end{figure}

\paragraph{Partial Rule Satisfaction in Conjunctive Concepts.}
Task 23 (GT: ``wooden floor living room'') exemplifies how annotators sometimes apply only partial matching to multi-component rules. As shown in Figure~\ref{fig:bongard-ow-23}, Images 4 and 5 in the positive set depict a dining area and a staircase respectively—neither of which are living rooms, despite having wooden floors. This creates fundamental ambiguity: should models learn that \textit{both} conditions must be satisfied (wooden floor AND living room), or that \textit{either} condition suffices? Such errors make it impossible to determine whether poor model performance reflects genuine conceptual limitations or simply confusion induced by contradictory training signals.

\begin{figure}[h]
    \centering
    \includegraphics[width=0.9\linewidth]{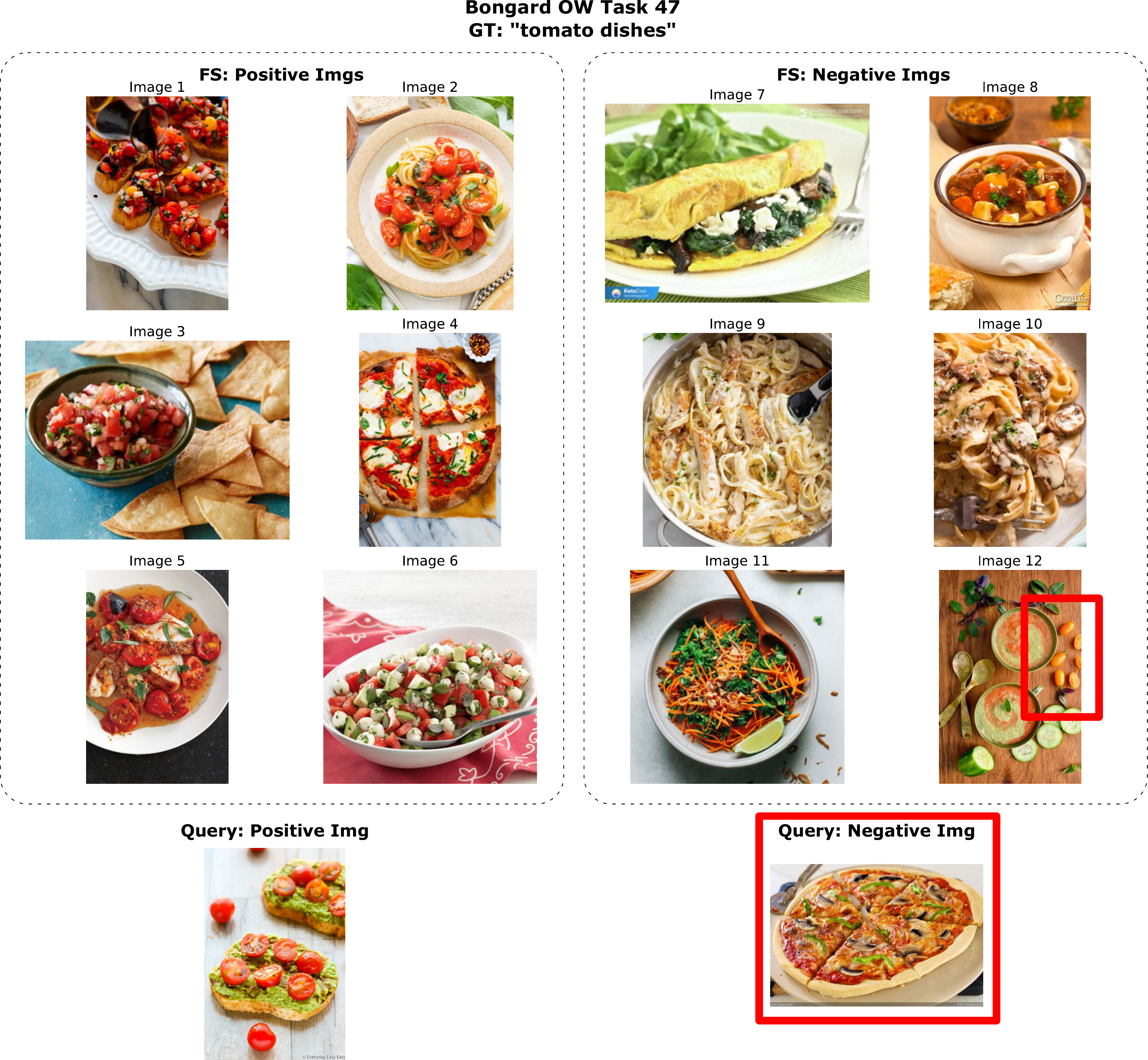}
    \caption{Annotation inconsistencies in Bongard-OW Task 47 (GT: ``tomato dishes''). The dataset contains contradictory labels: Image 12 (negative few-shot) clearly shows a tomato-based dish, while the negative query image shows a pizza with tomatoes despite Image 4 (positive few-shot) also being a pizza. These inconsistencies (highlighted in red) make the underlying rule ambiguous and evaluation unreliable.}
    \label{fig:bongard-ow-47}
\end{figure}

\paragraph{Contradictory Labels Within the Same Task.}
Task 47 (GT: ``tomato dishes'') contains multiple logical inconsistencies that render the rule unlearnable (Figure~\ref{fig:bongard-ow-47}). Image 12, labeled as a negative example, clearly shows a tomato-based dish. Meanwhile, Image 4 in the positive set shows a pizza, yet the negative query image also depicts a pizza with visible tomatoes. These contradictions make it impossible for any model—or human—to extract a coherent rule from the provided examples.

\subsection{Query Image Mislabeling}

\begin{figure}[h]
    \centering
    \includegraphics[width=0.9\linewidth]{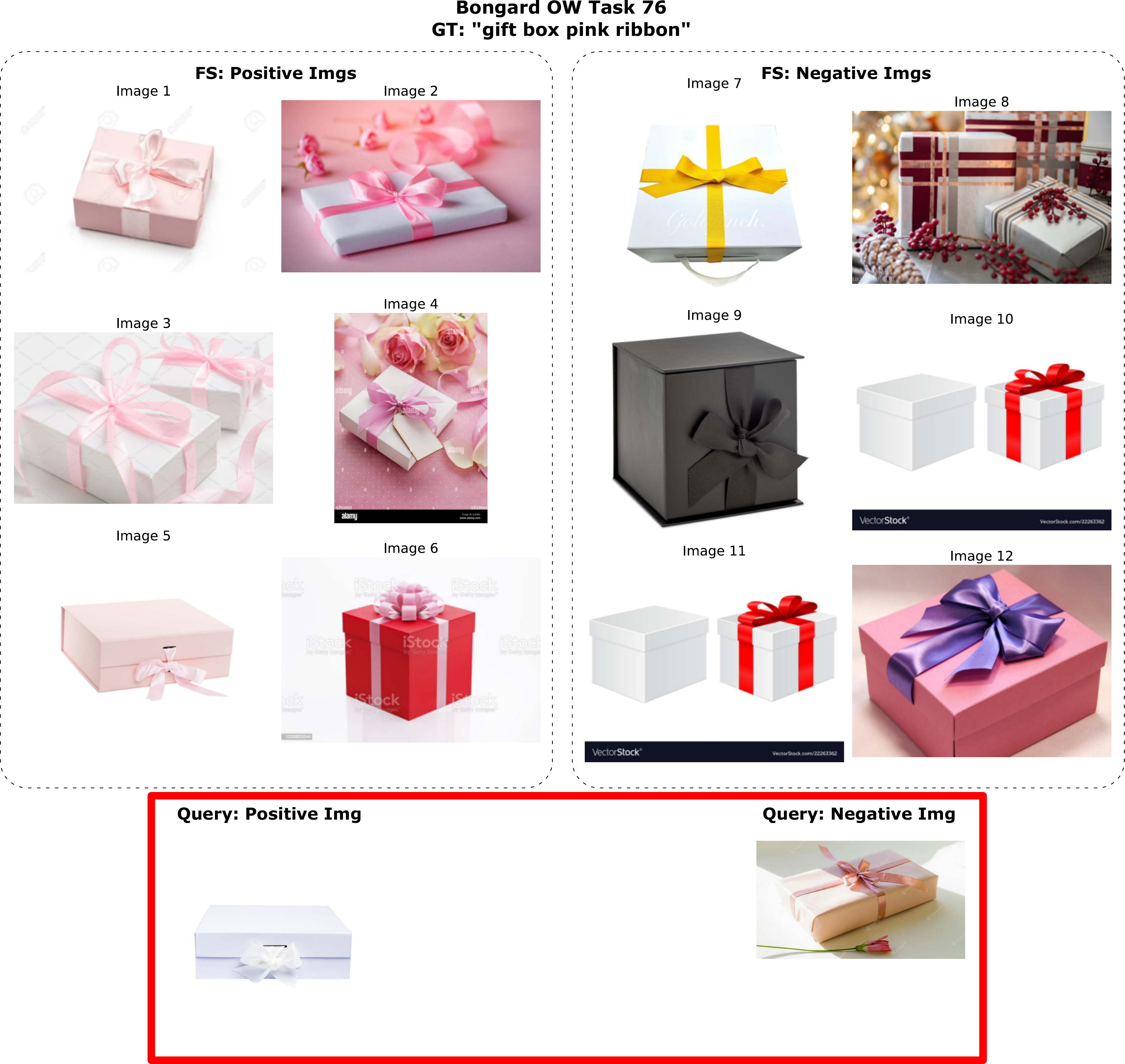}
    \caption{Query image mislabeling in Bongard-OW Task 76 (GT: ``gift box pink ribbon''). While the few-shot examples correctly distinguish pink ribbons (positive) from non-pink ribbons (negative), both query images (highlighted in red) contain white ribbons. This violates the ground-truth rule and renders the queries unanswerable based on the provided examples.}
    \label{fig:bongard-ow-76}
\end{figure}

\paragraph{Attribute Violations in Query Sets.}
Task 76 (GT: ``gift box pink ribbon'') demonstrates systematic query set errors (Figure~\ref{fig:bongard-ow-76}). While the few-shot examples correctly distinguish pink ribbons (positive) from other colors (negative), both query images contain white ribbons. Since neither query satisfies the ground-truth rule, their assigned labels are arbitrary, making evaluation metrics meaningless for this task.

\begin{figure}[h]
    \centering
    \includegraphics[width=0.9\linewidth]{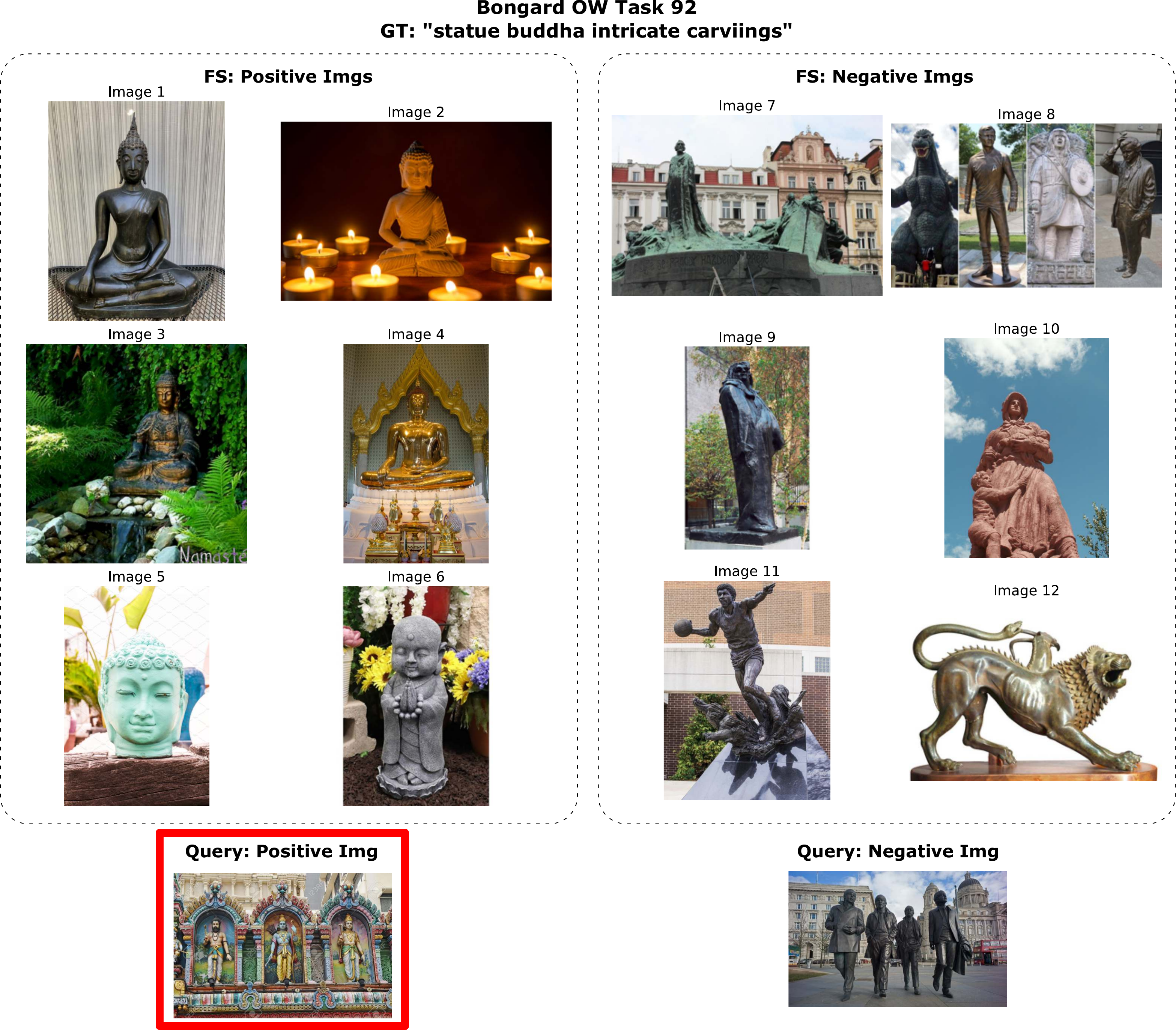}
    \caption{Annotation error in Bongard-OW Task 92 query set (GT: ``statue buddha intricate carvings''). While all positive few-shot images correctly show Buddha statues with intricate carvings, the positive query image (red box) contains Hindu deity statues instead. This categorical mismatch undermines the task's validity and prevents meaningful evaluation of rule generalization.}
    \label{fig:bongard-ow-92}
\end{figure}

\paragraph{Categorical Mismatches.}
Task 92 (GT: ``statue buddha intricate carvings'') reveals more severe errors where query images belong to entirely different conceptual categories (Figure~\ref{fig:bongard-ow-92}). The positive query depicts Hindu deity statues rather than Buddha statues—a fundamental categorical error, not merely a boundary case. Such mismatches indicate inadequate quality control in dataset construction and invalidate any assessment of model generalization.

\subsection{Implications for Performance Evaluation}

These examples are not isolated incidents but represent systematic issues that affect multiple tasks in the dataset. When ground-truth labels are internally contradictory or violate the stated rules, performance drops cannot be meaningfully interpreted. A model that fails on Task 47, for instance, may actually be learning the correct concept but struggling with the dataset's logical inconsistencies. Conversely, high accuracy on Task 76 may reflect memorization of spurious correlations rather than genuine rule understanding. These quality issues must be considered when interpreting any quantitative results on Bongard-OpenWorld.

\clearpage

\section{DSL used for VLP}\label{sec:dsl}
In \autoref{tab:dsl}, we present the DSL used in the experiments. While the functions are defined in a general form, they are adapted depending on the focus and characteristics of each dataset. For example, CLEVR-Hans contains synthetic objects that are not associated with actions, and therefore action related functions are unnecessary. Similarly, datasets with lower logical complexity do not require counting or numerical comparison operations, whereas Bongard-RWR and COCOLogic benefit from these additional functions due to their more demanding reasoning requirements.

\begin{table*}[h]
\centering
\footnotesize
\caption{Overview of all primitive functions of the DSL available for each dataset, together with their corresponding type signatures. The table groups functions by dataset and specifies the input and output types using arrow based type notation.}
\label{tab:dsl}
\begin{tabularx}{\textwidth}{l l X}
\toprule
\textbf{Dataset} & \textbf{Primitive} & \textbf{Type} \\
\midrule

\multirow{10}{*}{\textbf{Bongard-HOI}} 
& \texttt{get\_objects} & IMG $\to$ List(List(STRING)) \\
& \texttt{get\_actions} & IMG $\to$ List(List(STRING)) \\
& \texttt{exists\_object} & List(List(STRING)) $\to$ OBJECT $\to$ BOOL \\
& \texttt{exists\_object\_with\_property} & List(List(STRING)) $\to$ OBJECT $\to$ PROPERTY $\to$ BOOL \\
& \texttt{exists\_property} & List(List(STRING)) $\to$ PROPERTY $\to$ BOOL \\
& \texttt{exists\_action} & List(List(STRING)) $\to$ ACTION $\to$ BOOL \\
& \texttt{exists\_action\_with\_object} & List(List(STRING)) $\to$ ACTION $\to$ OBJECT $\to$ BOOL \\
& \texttt{and, or} & BOOL $\to$ BOOL $\to$ BOOL \\
& \texttt{not} & BOOL $\to$ BOOL \\

\midrule

\multirow{3}{*}{\textbf{Bongard-OW}} 
& Same as Bongard-HOI + \\
& \texttt{exists\_properties} & List(List(STRING)) $\to$ PROPERTY $\to$ PROPERTY $\to$ BOOL \\
& \texttt{exists\_object\_with\_properties} & List(List(STRING)) $\to$ OBJECT $\to$ PROPERTY $\to$ PROPERTY $\to$ BOOL \\

\midrule

\multirow{8}{*}{\textbf{Bongard-RWR}} 
& Same as Bongard-OW + \\
& \texttt{count\_object\_in\_img} & List(List(STRING)) \ensuremath{\rightarrow} OBJECT \ensuremath{\rightarrow} INT \\
& \texttt{count\_objects\_with\_property} & List(List(STRING)) \ensuremath{\rightarrow} PROPERTY \ensuremath{\rightarrow} INT \\
& \texttt{max\_objects\_of\_same\_type} & List(List(STRING)) \ensuremath{\rightarrow} INT \\
& \texttt{count\_all\_objects} & List(List(STRING)) \ensuremath{\rightarrow} INT \\
& \texttt{xor} & BOOL $\to$ BOOL $\to$ BOOL \\
& \texttt{gt?, eq?} & INT $\to$ INT $\to$ BOOL \\
& \texttt{0, 1, 2, 3, 4, 5, 6} & INT \\

\midrule

\multirow{8}{*}{\textbf{COCOLogic}} 
& Same as Bongard-HOI + & \\
& \texttt{count\_object\_in\_img} & List(List(STRING)) \ensuremath{\rightarrow} OBJECT \ensuremath{\rightarrow} INT \\
& \texttt{count\_objects\_with\_property} & List(List(STRING)) \ensuremath{\rightarrow} PROPERTY \ensuremath{\rightarrow} INT \\
& \texttt{max\_objects\_of\_same\_type} & List(List(STRING)) \ensuremath{\rightarrow} INT \\
& \texttt{count\_all\_objects} & List(List(STRING)) \ensuremath{\rightarrow} INT \\
& \texttt{xor} & BOOL $\to$ BOOL $\to$ BOOL \\
& \texttt{gt?, eq?} & INT $\to$ INT $\to$ BOOL \\
& \texttt{0, 1, 2, 3, 4, 5, 6} & INT \\

\midrule

\textbf{CLEVR-Hans3} & \multicolumn{2}{l}{Similar to Bongard-OW, but without action predicates.} \\

\bottomrule
\end{tabularx}
\end{table*}

\section{Prompts}\label{app:prompts}
In the following we provide the used prompts for the baseline \autoref{app:baseline_prompts}. For VLP we provide the prompts for symbol grounding \autoref{app:variable_discovery_prompts} as well as the VLM functions of VLP \autoref{app:vlm_functions}. Further we describe the VLM functions added in the course of RQ4 in \autoref{app:vlm_functions_size}. And finally the prompts for the ablation on program synthesis are provided in \autoref{app:structured_baseline_prompts}.

\subsection{Baseline Prompts}\label{app:baseline_prompts}
Prompts for the baseline used in the experiments. First, the VLM is prompted for a rule that separates the images of the problem, and then it is prompted to evaluate the rule on the test images.

\begin{llmprompt}{Baseline prompt for obtaining visual rule}
You are given \{n\} images. Each image depicts a scene with specific objects, interactions, and environments. \\
Your task is to determine the underlying concept that distinguishes the positive examples from the negative examples, based on the objects, their properties, and the actions occurring in each scene. \\

- The first \{m\} images are positive examples.\\
- The remaining \{o\} images are negative examples.\\

\#\# Task\\

Identify the rule that defines the positive examples:\\
- The rule must apply to all positive examples.\\
- The rule must not apply to any negative example.\\

\#\# Step-by-Step Process\\

1. Image Analysis:\\
- Carefully describe each image, noting objects, their attributes, and conceptual features (such as relationships, actions, or settings).\\

2. Rule Derivation:\\
- From your analysis, infer the rule that uniquely characterizes the positive examples.\\
- Confirm that the rule does not hold for the negative examples.\\

\#\# Final Answer Format\\

Provide your final answer in the following format:\\

```python\\
answer = \{\\
\quad'rule': '[RULE]',\\
\}\\
```\\

Ensure that the rule is clearly defined and concise.
\end{llmprompt}

\begin{llmprompt}{Baseline prompt for evaluating direct result}
Given the rule '\{response\}', determine if the image follows the rule or not. Answer with 'Yes' or 'No', nothing else.
\end{llmprompt}

\subsection{Symbol Grounding}\label{app:variable_discovery_prompts}
In the following the prompts used during symbol grounding are provided. The VLM receives the task images together with the prompt instructions to identify relevant \textit{objects}, \textit{properties} and \textit{actions}.

For the amount of symbols requested, we use the parameters listed in \autoref{tab:dataset_variables} across all experiments.

\begin{table}[h]
    \centering
    \caption{\textbf{Variables used per dataset in program search.} Number of objects, properties, and actions requested when generating programs for each dataset.}
    \label{tab:dataset_variables}
    \vskip 0.4em
    \small
    \begin{tabular}{lccc}
    \toprule
    \textbf{Dataset} & \textbf{\#Objects} & \textbf{\#Properties} & \textbf{\#Actions} \\
    \midrule
    Bongard-HOI & $10$ & $5$  & $10$ \\
    Bongard-OW  & $10$ & $10$ & $3$ \\
    Bongard-RWR & $10$ & $10$ & $5$ \\
    COCOLogic   & $10$ & $10$ & $3$ \\
    CLEVR-Hans3 & $10$ & $10$ & $0$ \\
    \bottomrule
    \end{tabular}
\end{table}

\begin{llmprompt}{Objects}
You are analyzing images to identify notable objects. Focus on clearly identifiable, semantically meaningful objects that appear across the image set, especially those present in some images but not others. Objects include persons, animals, and things. Avoid minor details like shadows or textures. Use specific, descriptive names (e.g., "bicycle" not "vehicle"). Return exactly \{n\} objects in a Python list.\\

Answer format:\\
```python\\
objects = [...]\\
```\\
No comments or explanations. If no objects found, return [].
\end{llmprompt}

\begin{llmprompt}{Properties}
\# Property Discovery Task\\

You are analyzing a set of images to identify important properties that describe objects in the image set.

\#\# Objective\\

Discover **notable properties** that characterize objects in the images. Focus on properties that meaningfully distinguish or describe objects (visual attributes, spatial relationships, geometric features, states).\\

\#\# Instructions\\

1. **Examine all images carefully** - Look for properties that apply to the relevant objects across the image set\\
2. **Identify important properties** - Focus on significant, clearly observable properties that meaningfully describe objects\\
3. **Consider property variation** - Properties that vary across images or objects may be particularly noteworthy\\
4. **Prioritize meaningful properties** - Choose properties that help distinguish or characterize objects (e.g., color, size, position, orientation, state)\\
5. **Return exactly {n} properties** - If fewer notable properties exist, return as many as available\\
6. **Use descriptive names** - Name properties clearly and specifically (e.g., "red" rather than "colored", "horizontal" rather than "oriented")\\

\#\# Relevant Objects\\
The objects to consider are: \{objects\}\\

\#\# Property Categories\\
- **Visual attributes**: color, texture, pattern, brightness\\
- **Spatial properties**: position (left/right, top/bottom, center), proximity, distance\\
- **Geometric attributes**: size, shape, orientation, symmetry\\
- **States**: filled/outlined, open/closed, active/inactive\\

\#\# Output Requirements\\
- Return a Python list assigned to variable `properties`\\
- Include only the Python code, no explanations or comments\\
- If no notable properties are found, return an empty list `[]`\\
- Use clear, specific property names (e.g., "blue", "large", "leftmost", "vertical")\\
- Be general (e.g., if there's a yellow triangle, use "yellow" not "yellow triangle")\\

\#\# Example Format\\
```python\\
properties = ["red", "large", "horizontal", "outlined", "centered"]\\
```
\end{llmprompt}

\begin{llmprompt}{Actions}
\# Action Discovery Task\\

You are analyzing a set of images to identify important actions performed by objects in the image set.\\

\#\# Objective\\

Discover **notable actions** that characterize objects in the images. Focus on actions that meaningfully distinguish or describe what objects are doing (movements, behaviors, states of activity).\\

\#\# Instructions\\

1. **Examine all images carefully** - Look for actions that apply to the relevant objects across the image set\\
2. **Identify important actions** - Focus on significant, clearly observable actions that meaningfully describe what objects are doing\\
3. **Consider action variation** - Actions that vary across images or objects may be particularly noteworthy (contrasting actions)\\
4. **Prioritize meaningful actions** - Choose actions that help distinguish or characterize object behaviors (e.g., running, jumping, standing, flying)\\
5. **Return exactly {n} actions** - If fewer notable actions exist, return as many as available\\
6. **Use descriptive names** - Name actions clearly and specifically (e.g., "running" rather than "moving", "sitting" rather than "positioned")\\

\#\# Relevant Objects\\

The objects to consider are: \{objects\}\\

\#\# Action Categories\\

- **Movement actions**: walking, running, jumping, flying, rolling\\
- **Positional actions**: standing, sitting, lying, hanging\\
- **Interactive actions**: holding, pushing, pulling, touching\\
- **State actions**: opening, closing, rotating, tilting\\

\#\# Output Requirements\\

- Return a Python list assigned to variable `actions`\\
- Include only the Python code, no explanations or comments\\
- If no notable actions are found, return an empty list `[]`\\
- Use clear, specific action names (e.g., "jumping", "sitting", "rotating", "falling")\\

\#\# Example Format\\
```python\\
actions = ["running", "jumping", "standing", "flying", "sitting"]\\
```
\end{llmprompt}

\subsection{VLM Functions of the DSL}\label{app:vlm_functions}
In the following the two VLM functions used for VLP are presented. These are executed on individual images to obtain a structured representation of the objects and actions in the image.

\begin{llmprompt}{get\_objects - VLM function for obtaining object-property representation}
{\scriptsize
\#\# Task\\
Identify objects and their properties from the image using only the provided lists.\\

**Objects:** \{objects\}\\
**Properties:** \{properties\}\\

\#\# Rules\\
1. Only use objects/properties from the provided lists\\
2. Return empty list if no valid objects found\\
3. No explanations or additional text\\

\#\# Output Format\\
```python\\
objects = [ 

\quad['object\_name', 'property1', 'property2', ...],
    
\quad['object\_name', 'property1'], 
    
    ... 
    
] 
```\\

**If no valid objects:** `objects = [[]]` \\

\#\# Examples\\

**Example 1**\\
- Objects: ["car", "person", "tree"]\\
- Properties: ["red", "tall", "small", "standing"]\\
- Image: Red car under tall tree with small standing person\\

```python\\
objects = [

\quad['car', 'red'],
    
\quad['tree', 'tall'], 
    
\quad['person', 'standing', 'small']
    
]\\
```\\

**Example 2**\\
- Objects: ["dog", "ball", "book", "chair"] \\
- Properties: ["blue", "sitting", "round"] \\
- Image: Dog sitting by round ball and blue chair \\

```python\\
objects = [

\quad['dog', 'sitting'],
    
\quad['ball', 'round'],
    
\quad['chair', 'blue']
    
]\\
```\\

**Example 3**\\
- Objects: ["bicycle", "lamp", "table", "cup"]\\
- Properties: ["green", "broken", "wooden", "white"]\\
- Image: Table with laptop and cup\\

```python\\
objects = [[]]\\
```\\
*Note: Even though 'table' and 'cup' are in the objects list and visible in the image, neither has properties from the provided list, so no valid object-property combinations exist*\\

**Analyze the image now:**}
\end{llmprompt}

\begin{llmprompt}{get\_actions - VLM function for obtaining action-object representation}
{\scriptsize
\#\# Task\\
Identify actions occurring in the image using only the provided lists.\\

**Actions:** {actions} \\
**Objects:** {objects} \\

\#\# Rules\\
1. Only use actions/objects from the provided lists\\
2. Only detect actions that are actually happening in the image\\
3. Do not include actions from the list if they are not occurring in the image\\
4. If an action involves an object, include the object name\\
5. Return empty list if no valid actions found\\
6. No explanations or additional text\\

\#\# Output Format\\
```python\\
actions = [

\quad['action\_name1'],
    
\quad['action\_name2', 'object\_name2'],
    
\quad['action\_name2', 'object\_name1', 'object\_name2'],
    
\quad...

]\\
```\\

**If no valid actions:** `actions = [[]]`\\

\#\# Examples\\

**Example 1**\\
- Actions: ["running", "jumping", "sitting", "dancing"]\\
- Objects: ["chair", "ball", "person", "table"]\\
- Image: Person sitting on chair\\

```python\\
actions = [

\quad['sitting', 'person', 'chair']

]\\
```\\
*Note: 'running', 'jumping', and 'dancing' are in the actions list but not happening in the image, so they're excluded*\\

**Example 2**\\
- Actions: ["throwing", "catching", "walking", "reading", "sleeping"]\\
- Objects: ["ball", "book", "dog", "frisbee"]\\
- Image: Person throwing a ball while dog is walking\\

```python\\
actions = [

\quad['throwing', 'person', 'ball'],

\quad['walking', 'dog']

]\\
```\\
*Note: 'catching', 'reading', and 'sleeping' are in the actions list but not occurring in the image, so they're excluded*\\

**Example 3**\\
- Actions: ["swimming", "flying", "cooking"]\\
- Objects: ["pool", "bird", "kitchen"]\\
- Image: Person eating at a restaurant\\

```python\\
actions = [[]]\\
```\\
*Note: Even though actions are happening in the image, none match the provided actions list, so no valid actions exist*\\

**Analyze the image now:**}
\end{llmprompt}

\subsection{VLM Functions for property size}\label{app:vlm_functions_size}
In the course of (RQ4) we add VLM functions to the DSL that explicitly ask the VLM for objects of small and large size. These are listed in the following.

\begin{llmprompt}{exists\_object\_small\_in\_img}
Does the image contain any '\{obj\}' that is relatively small in size compared to the other objects? Answer with 'YES' or 'NO'.
\end{llmprompt}

\begin{llmprompt}{exists\_object\_large\_in\_img}
Does the image contain any '\{obj\}' that is relatively large in size compared to the other objects? Answer with 'YES' or 'NO'.
\end{llmprompt}

\begin{llmprompt}{exists\_object\_with\_property\_small\_in\_img}
Does the image contain any '\{obj\}' with the property '\{prop\}' that is relatively small in size compared to the other objects? Answer with 'YES' or 'NO'.
\end{llmprompt}

\begin{llmprompt}{exists\_object\_with\_property\_large\_in\_img}
Does the image contain any '\{obj\}' with the property '\{prop\}' that is relatively large in size compared to the other objects? Answer with 'YES' or 'NO'.
\end{llmprompt}

\subsection{Structure-based baseline}\label{app:structured_baseline_prompts}
For the ablation study on program synthesis in \autoref{app:reasoning_ablation}, we implement a structure-based approach that uses the VLM functions defined in the DSL (see above) to extract symbolic representations of the input images. These representations are then passed back to the VLM, which reasons over them to induce a rule. Below we provide the prompt used for this procedure, along with the prompt applied during evaluation when test images are assessed based on their structure-based representations.

\begin{llmprompt}{Structure-based rule induction}
{\scriptsize
\# Task: Concept Identification from Structured Image Representations\\

You will be given a sequence of structured image representations, where each image is labeled as either a **positive** or **negative** example of an underlying concept.\\

\#\# Structure of Each Image Representation\\

\#\#\# Objects\\
A list of objects, where each object is described together with its properties.\\

**Format:**\\
```

[[object\_name, property1, property2, ...], [object\_name, property1, ...], ...]

```

**Example:**\\
```

[[dog, brown, large], [cat, small, cute], [ball, red, round]]

```

\#\#\# Actions\\
A list of actions occurring in the image, including the entities involved.\\

**Format:**\\
```

[[action, participant1, participant2, ...], [action, participant], ...]

```

**Example:**\\
```

[[playing, dog, cat], [shining, sun], [rolling, ball]]

```

\#\# Your Objective

Identify the rule that defines the positive examples:\\
- The rule must apply to all positive examples.\\
- The rule must not apply to any negative example.\\

\#\# Step-by-Step Process

1. Image Analysis:\\
- Analyze the objects, their properties, and the actions across all examples to identify the concept that distinguishes positive examples from negative ones.\\

2. Rule Derivation:\\
- From your analysis, infer the rule that uniquely characterizes the positive examples.\\
- Confirm that the rule does not hold for the negative examples.\\

\#\# Structured Image Representations

Image: 1\\
Objects: \{obj\_repr\}\\
Actions: \{action\_repr\}\\
Label: 'Positive'\\

...\\

Image: \{n\}\\
Objects: \{obj\_repr\}\\
Actions: \{action\_repr\}\\
Label: 'Negative'\\

\#\# Final Answer Format

Provide your final answer in the following format:

```python\\
answer = \{\\
\quad'rule': '[RULE]',\\
\}\\
```\\

Ensure that the rule is clearly defined and concise.
}
\end{llmprompt}

\begin{llmprompt}{Structure-based evaluation}
\# Task: Classify Image based on Structured Image Representations\\

You are given a structured image representation.\\

\#\# Structure of Each Image Representation\\

\#\#\# Objects\\
A list of objects, where each object is described together with its properties.\\

**Format:**\\
```

[[object\_name, property1, property2, ...], [object\_name, property1, ...], ...]

```

**Example:**\\
```

[[dog, brown, large], [cat, small, cute], [ball, red, round]]

```

\#\#\# Actions\\
A list of actions occurring in the image, including the entities involved.\\

**Format:**\\
```

[[action, participant1, participant2, ...], [action, participant], ...]

```

**Example:**\\
```

[[playing, dog, cat], [shining, sun], [rolling, ball]]

```

\#\# Your Objective\\
Given the rule '\{rule\}', determine if the image follows the rule or not. Answer with 'Yes' or 'No', nothing else.\\

\#\# Structured Image Representations

\{representation\}
\end{llmprompt}